\documentclass[11pt]{article}

\usepackage[final]{acl}
\usepackage{times}
\usepackage{latexsym}
\usepackage{makecell}

\usepackage{tikz}
\usetikzlibrary{arrows.meta,positioning}

\usepackage{pgfplots}
\pgfplotsset{compat=1.18}
\usepackage{booktabs}
\usepackage{multirow}
\usepackage{xcolor}
\usepackage{colortbl}
\newcommand{\best}[1]{\textbf{\textcolor{blue!70!black}{#1}}} 
\newcommand{\RankOne}[1]{\textbf{\textcolor{yellow!50!orange}{#1}}}
\newcommand{\RankTwo}[1]{\textbf{\textcolor{cyan}{#1}}}
\newcommand{\RankThree}[1]{\textbf{\textcolor{lime!80!black}{#1}}}
\newcommand{\strong}[1]{\textbf{\textcolor{red!70!black}{#1}}} 

\newcommand{\ContamV}[1]{\multicolumn{1}{!{\color{red!75!black}\vrule width 1.2pt}r}{#1}}

\usepackage[most]{tcolorbox}
\usepackage{booktabs}
\usepackage{enumitem}

\usepackage{booktabs}
\usepackage{pifont}

\usepackage{booktabs}
\usepackage{tabularx}
\usepackage{array}

\newcolumntype{Y}{>{\raggedright\arraybackslash}X}

\usepackage[T1]{fontenc}

\usepackage[utf8]{inputenc}

\usepackage{microtype}

\usepackage{inconsolata}

\usepackage{graphicx}
\usepackage{caption} 
\newcounter{savedtablecounter}

\usepackage{marvosym}

\usepackage{fontawesome5}

\newcommand{\medal}[1]{\raisebox{-0.25\height}{\includegraphics[height=1.15em]{#1}}}

\newcommand{\WinnerMedal}{\medal{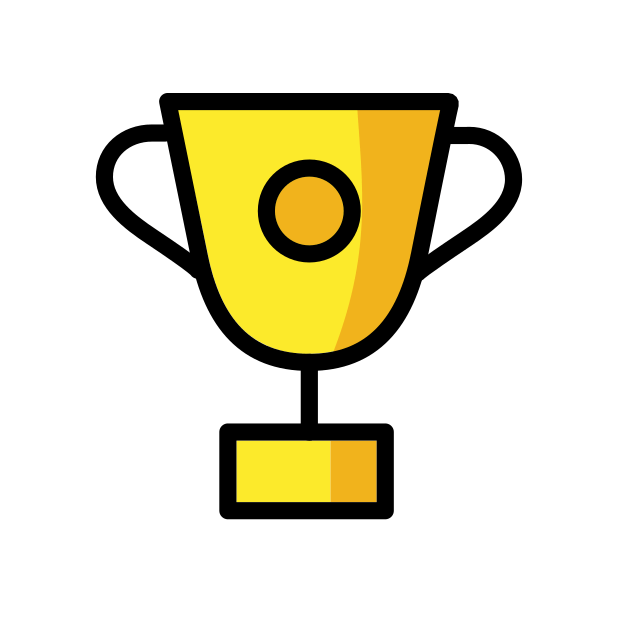}}
\newcommand{\GoldMedal}{\medal{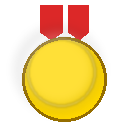}}
\newcommand{\SilverMedal}{\medal{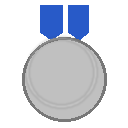}}
\newcommand{\BronzeMedal}{\medal{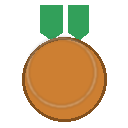}}

\newcommand{\Winner}[1]{\WinnerMedal\,#1}
\newcommand{\Gold}[1]{\GoldMedal\,#1}
\newcommand{\Silver}[1]{\SilverMedal\,#1}
\newcommand{\Bronze}[1]{\BronzeMedal\,#1}

\title{\texorpdfstring{\raisebox{-0.8ex}{\includegraphics[height=1.8em]{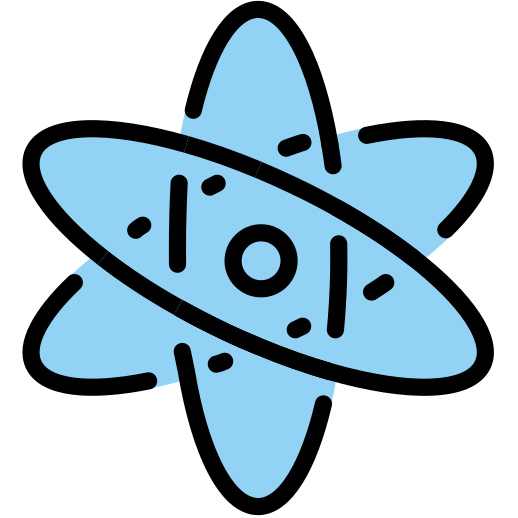}}}{} ScienceArena: Benchmarking LLMs on Latest Scientific Olympiad Competitions}

\author{
 \textbf{Guangxiang Zhao\textsuperscript{1,}\textsuperscript{\faCrown}},
 \textbf{Qilong Shi\textsuperscript{2,}\textsuperscript{\faCrown}},
 \textbf{Xusen Xiao\textsuperscript{3,}\textsuperscript{\faCrown}},
 \textbf{Wenpu Liu\textsuperscript{4}},
 \textbf{Yaoming Li\textsuperscript{4,}\textsuperscript{\Letter}},\\
 \textbf{Linfeng Hao\textsuperscript{4}},
 \textbf{Shuyang Hou\textsuperscript{4}},
 \textbf{Zijian Guo\textsuperscript{4}},
 \textbf{Xinrui Zhang\textsuperscript{4}},
 \textbf{Yuntian Zhao\textsuperscript{4}},\\
 \textbf{Zhengyang Wang\textsuperscript{4}},
 \textbf{Wenrui Liu\textsuperscript{4}},
 \textbf{Yuhan Wu\textsuperscript{4}},
 \textbf{Tong Yang\textsuperscript{4,}\textsuperscript{\Letter}},
 \textbf{Lin Sun\textsuperscript{1,}\textsuperscript{\Letter}},
 \textbf{Xiangzheng Zhang\textsuperscript{1}}
\\
 \textsuperscript{1}Qiyuan Tech,
 \textsuperscript{2}Tsinghua University,
 \textsuperscript{3}The University of Hong Kong,
 \textsuperscript{4}Peking University
\\
 \small{
   \Letter\ \textbf{Correspondence:} \href{mailto:yibachenggong@gmail.com}{yibachenggong@gmail.com}, \href{mailto:yangtong@pku.edu.cn}{yangtong@pku.edu.cn}, \href{sunlin1@360.cn}{sunlin1@360.cn}
 }
}

\begin{document}
\renewcommand{\thefootnote}{\fnsymbol{footnote}}
\maketitle
\begingroup
\renewcommand{\thefootnote}{\faCrown}
\footnotetext{\ The first three authors contributed equally.}
\endgroup
\begin{abstract}
Benchmark saturation and data contamination increasingly obscure genuine scientific reasoning in frontier LLMs. We introduce \textsc{ScienceArena}, an olympiad-style benchmark from thirteen public science competitions in physics, chemistry, and biology, including IPhO and IChO 2025--2026, IBO 2023, USAPhO 2026, and USNCO 2025. Its open-ended, multi-step problems use process-credit rubrics, making faithful scoring difficult. We build ScienceArena through an expert-audited digitization pipeline that converts official exams, figures, solutions, and rubrics into structured items verified by olympiad medalists. To scale evaluation beyond costly human grading, we calibrate LLM-as-judge against medalist ground truth on archived answers from five models across IPhO and IChO; two strong judges stay within one point of expert total scores. Medalist notes show that failures often stem from visual grounding, structure fidelity, and global problem control rather than missing terminology. Evaluating fourteen recent LLMs with interleaved solving, we find that top models obtain medal-equivalent rubric scores on several public international exams, while chemistry and long-horizon consistency remain key bottlenecks. We provide an interactive \href{https://science-arena.onrender.com/}{demo}\footnote{\ Demo: \url{https://science-arena.onrender.com/}.}.
\end{abstract}

\section{Introduction}
Scientific reasoning has become a central target for frontier LLMs, yet evaluation is at an impasse.
Popular benchmarks are increasingly saturated and vulnerable to contamination \citep{hendrycks2021mmlu,li2023contamination,rein2023gpqa,white2024livebench,phan2025hle}.
Recent live or frontier benchmarks respond by using time-sensitive sources, contest problems, or newly authored expert questions \citep{jain2024livecodebench,glazer2024frontiermath,balunovic2025matharena}, but scientific reasoning remains difficult to measure when answers are long, visual, and partially correct.
At the same time, many science evaluations remain confined to short-answer or multiple-choice formats, which cannot fully test the abilities needed in real scientific problem solving: long derivations, diagram and structure interpretation, multi-step consistency, and partial credit under explicit rubrics \citep{wang2023scibench,lu2023mathvista,yue2023mmmu,he2024olympiadbench,lu2022scienceqa,laurent2024labbench,mirza2025chembench}.

Science olympiad competitions are a natural high-resolution testbed.
They are written by domain experts, released publicly after the contest, and accompanied by official answers, detailed solutions, and scoring schemes.
They also provide interpretable human reference points, including gold-medal thresholds and winner scores.
However, turning olympiad exams into a reliable LLM benchmark is not automatic.
The source materials are PDFs with formulas, figures, tables, and chemical structures; more importantly, faithful scoring requires stepwise rubrics and expert judgment rather than simple exact matching.

We present \textsc{ScienceArena}, an olympiad-style benchmark and evaluation protocol for scientific reasoning.
The suite contains thirteen physics, chemistry, and biology contest-year datasets released in 2023--2026. An archived five-model subset covers IPhO 2025, IChO 2025, and IBO 2023; the full evaluation also includes international exams such as IPhO and IChO 2026 and national-level exams such as USAPhO and USNCO.
We standardize the data construction pipeline: official public materials are collected, converted into structured JSON with OCR and multimodal parsing, and then audited by olympiad medalists for problem text, figure descriptions, answers, detailed solutions, rubrics, and point totals.
This preserves the original exam structure while making the problems readable to text-based and multimodal LLM APIs.

\begin{table*}[t]
\centering
\scriptsize
\setlength{\tabcolsep}{5pt}
\renewcommand{\arraystretch}{0.95}
\begin{tabularx}{\textwidth}{p{0.19\textwidth} p{0.23\textwidth} Y}
\toprule
\textbf{Benchmark} & \textbf{Main signal} & \textbf{ScienceArena distinction} \\
\midrule
Chatbot Arena; Arena-Hard
& Human or judge preference
& Tests verifiable scientific correctness rather than broad helpfulness. \\
\addlinespace[0.12em]
OlympicArena/OlympiadBench
& Cross-discipline hard reasoning
& Keeps recent science exams, release dates, official rubrics, and medalist audits. \\
\addlinespace[0.12em]
MathArena
& Fresh math answers/proofs
& Adds physics, chemistry, biology, diagrams, structures, and science process credit. \\
\addlinespace[0.12em]
HiPhO/PhyArena/PhysicsArena
& Physics olympiad or skill axes
& Extends olympiad-style evaluation across three sciences with one unified protocol. \\
\addlinespace[0.12em]
SUPERChem; BABE
& Domain research or expert tasks
& Uses official competitions with comparable scoring scales and medal references. \\
\addlinespace[0.12em]
SciArena
& Researcher preference
& Scores against official answers and rubrics, then analyzes medalist grading notes. \\
\midrule
\textsc{ScienceArena}
& Release-aware olympiad scoring
& Combines official credit rubrics, calibrated judges, medals, and expert diagnostics. \\
\bottomrule
\end{tabularx}
\vspace{-0.1cm}
\caption{Positioning of \textsc{ScienceArena} among representative Arena-style and domain-specific evaluations. The comparison highlights the niche targeted by this paper: release-aware science olympiad evaluation with official process-credit rubrics and medalist-audited diagnostics.}
\label{tab:arena-positioning}
\vspace{-0.5cm}
\end{table*}

Table~\ref{tab:arena-positioning} positions \textsc{ScienceArena} relative to recent Arena-style and domain-specific evaluations.
General-purpose arenas such as Chatbot Arena and Arena-Hard emphasize human or judge-model preferences over open-ended assistant behavior \citep{chiang2024chatbotarena,li2024arenahard}.
Olympiad and competition benchmarks provide harder reasoning tasks, including OlympicArena, OlympiadBench, and MathArena \citep{huang2024olympicarena,he2024olympiadbench,balunovic2025matharena}.
The closest physics-only line is HiPhO/PhyArena, which uses latest physics olympiad exams and official medal thresholds \citep{yu2025hipho}; other domain benchmarks decompose physics, chemistry, biology, or scientific literature reasoning with different task sources and scoring signals \citep{dai2025physicsarena,zhao2025superchem,zhou2026babe,zhao2025sciarena}.
Our focus is complementary: recent public olympiad exams across physics, chemistry, and biology, official process-credit rubrics, medal-based human reference points, medalist-audited construction, and expert-calibrated automated grading.

A second challenge is evaluation cost.
Human olympiad experts provide the most reliable ground truth, but repeatedly asking them to grade every new model is impractical.
We therefore develop an expert-calibrated LLM-as-judge system.
Using human medalist scores on IPhO/IChO 2025 as ground truth, we calibrate the full judge system, including rubric packaging, structured scoring, score validation, and aggregation, on archived answers from five models.
The resulting judge system follows recent rubric-based LLM evaluation practice \citep{zheng2023judging,rao2026autorubric} and matches expert total scores within $\pm 1$ point across the two exams, making automated scoring practical while retaining an explicit human validity check.

Finally, we examine how LLMs should solve long olympiad problems.
One-shot prompting asks for an entire multi-part solution at once, while interleaved prompting presents subparts sequentially and preserves prior context.
We do not claim interleaving as a new method; rather, we test this natural prompting choice and find that it generally improves performance and better resembles how users interact with LLMs on difficult problems.
We therefore use interleaved prompting for the main evaluation.
Across fourteen recent LLMs, the strongest systems now obtain medal-equivalent rubric scores on several public international exams.
The gains are uneven: biology appears closest to saturation, whereas chemistry remains challenging because of molecular structures, stereochemistry, spectra, and rubric-sensitive drawings; physics still exposes failures in visual grounding and long-horizon coherence.
Beyond aggregate scores, we study the written comments produced by olympiad medalist graders.
These notes are unusually diagnostic: they distinguish fluent but physically misgrounded derivations from correct problem models, separate chemically plausible prose from exact structure commitments, and expose when partial credit depends on an explicit intermediate rather than a vague explanation.
We use these comments to build the subfield and ability-boundary analysis in Section~\ref{sec:expert-insights} and Appendix~\ref{app:expert_notes}.

\begin{figure*}[htbp]
\centering
\setlength{\abovecaptionskip}{0.1cm}
\includegraphics[width=\linewidth]{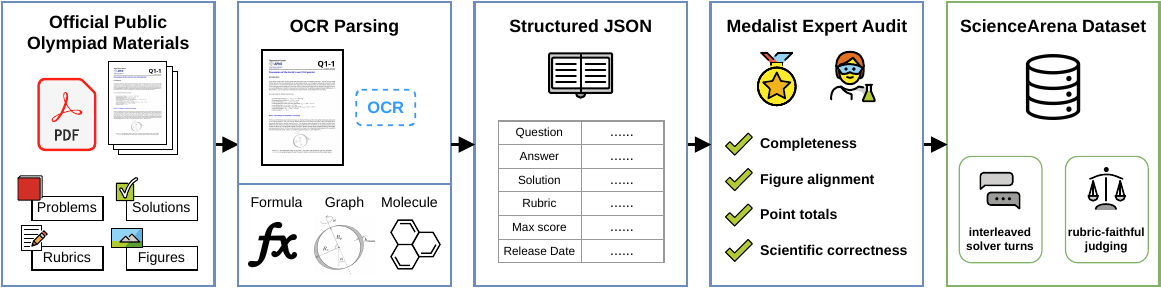}
\caption{\textsc{ScienceArena} benchmark construction pipeline.}
\label{fig:benchmark-pipeline}
\vspace{-0.5cm}
\end{figure*}

\paragraph{Contributions.}
We make four contributions:
\begin{itemize}[leftmargin=0.5em,itemsep=0.1em]
    \item We build \textsc{ScienceArena}, an expert-audited benchmark of thirteen science olympiad competitions with questions, figures, official answers, detailed solutions, and scoring rubrics.
    \item We define a reusable pipeline for converting official olympiad PDFs into structured LLM-readable benchmark items.
    \item We develop an expert-calibrated LLM-as-judge system that closely matches olympiad-medalist grading on IPhO \& IChO calibration experiments.
    \item We compare one-shot and interleaved prompting, evaluate fourteen recent LLMs under the stronger protocol, and distill comments from human experts into diagnostics of persistent domain-specific bottlenecks.
\end{itemize}

\section{The ScienceArena Framework}\label{sec:2}
\textsc{ScienceArena} is designed around two constraints that are often absent from standard LLM benchmarks.
First, the source tasks must remain fresh, public, and scientifically meaningful.
Second, grading must respect olympiad rubrics: a wrong final value may still receive credit for a correct setup, while a correct-looking answer may receive little credit if the derivation or diagram is unsupported.
This section describes the benchmark construction pipeline, the calibrated judge system, and the solving protocols used for the experiments.


\subsection{Benchmark Construction}
\textsc{ScienceArena} currently contains thirteen public olympiad-style competitions across physics, chemistry, and biology (Table~\ref{tab:science_arena_sources}).
International competitions provide medal thresholds; national-level competitions expand topic diversity and reduce dependence on one exam style.
For every benchmark, we record the public release date and preserve the original hierarchy: each multi-part problem becomes linked turns with text, figure context, official answer, rubric, and maximum score.
This keeps ru-brics intact while supporting automated evaluation.

\begin{table}[t]
\centering
\footnotesize
\setlength{\tabcolsep}{3pt}
\renewcommand{\arraystretch}{1.08}
\resizebox{\columnwidth}{!}{
\begin{tabular}{l l l c r}
\toprule
\textbf{Benchmark} & \textbf{Subject} & \textbf{Level} & \textbf{Rel.} & \textbf{Scale} \\
\midrule
IPhO 2025 & Physics & International & 25-07 & 30 \\
IChO 2025 & Chemistry & International & 25-07 & 60 \\
IBO 2023 & Biology & International & 25-08 & 453 \\
IPhO 2026 & Physics & International & 26-07 & 30 \\
IChO 2026 & Chemistry & International & 26-07 & 60 \\
APhO 2025 & Physics & Continental & 25-05 & 30 \\
EuPhO 2025 & Physics & Continental & 25-06 & 30 \\
NBPhO 2026 & Physics & National & 26-04 & 72 \\
USAPhO 2026 & Physics & National & 26-05 & 150 \\
CPhO 2025 & Physics & National & 25-10 & 320 \\
INChO 2026 & Chemistry & National & 26-01 & 106 \\
USNCO 2025 & Chemistry & National & 25-06 & 100 \\
CChO 2025 & Chemistry & National & 25-10 & 100 \\
\bottomrule
\end{tabular}
}
\vspace{-0.2cm}
\caption{\textsc{ScienceArena} benchmark sources. Release dates use YY-MM. ``Scale'' is the full score; rubric status and source caveats are retained in the released metadata.}
\label{tab:science_arena_sources}
\vspace{-0.5cm}
\end{table}

\paragraph{Digitization pipeline.}
Olympiad materials are PDFs with equations, plots, molecule drawings, apparatus diagrams, answer images, and marking schemes, so we convert them through a five-stage pipeline.
A curator first collects official public problems, figures, answer keys, solutions, and marking schemes; pages are then rendered and parsed with OCR/multimodal models.
The parsed content is normalized into fields such as \texttt{question\_text}, \texttt{official\_answer}, \texttt{official\_detailed\_solution}, \texttt{official\_rubric}, \texttt{max\_score}, \texttt{release\_date}, and source provenance.
Subject medalists audit completeness, figure alignment, answer fidelity, and point totals; for IPhO and IChO, the ground-truth graders are listed in Table~\ref{tab:human-graders}.
The final export is a unified interleaved dataset with one row per subproblem turn.
Figure~\ref{fig:benchmark-pipeline} summarizes the intended five-stage construction schematic.

The audit trail is important because the three sciences stress different failure modes.
Physics often requires plot reading and symbolic carry-over; chemistry depends on reaction schemes, stereochemistry, spectra, and crystal structures; biology is more often objective-keyed but can involve long passages and diagrams.
Preserving sources, parsed text, image-derived descriptions, and audit notes makes these differences inspectable rather than hiding them behind one score.

\begin{table}[ht]
\vspace{-0.1cm}
\centering
\footnotesize
\setlength{\tabcolsep}{4pt}
\renewcommand{\arraystretch}{1.15}
\begin{tabular}{l c c}
\toprule
\textbf{Track} & \textbf{Expert 1} & \textbf{Expert 2} \\
\midrule
Physics &
\makecell{Physics Olympiad\\\RankOne{gold medalist}\\\strong{world top 10}} &
\makecell{Physics Olympiad\\\RankOne{gold medalist}\\\strong{world top 10}} \\
\midrule
Chemistry &
\makecell{Chemistry Olympiad\\\RankOne{gold medalist}\\\strong{national top 30}} &
\makecell{Chemistry Olympiad\\\RankOne{gold medalist}\\\strong{national top 30}} \\
\midrule
Biology &
\makecell{Biology Olympiad\\\RankOne{gold medalist}\\\strong{national top 30}} &
\makecell{Biology Olympiad\\\RankOne{gold medalist}\\\strong{national top 30}} \\
\bottomrule
\end{tabular}
\vspace{-0.1cm}
\caption{Human experts recruited for ground-truth grading. The two experts for each exam jointly graded model answers and resolved rubric-level scores.}
\label{tab:human-graders}
\vspace{-0.3cm}
\end{table}

\subsection{Expert-Calibrated LLM-as-Judge System}
\label{sec:judge-system}
Human olympiad experts are the strongest graders for open-ended science problems, but they cannot regrade every model update.
We therefore use medalist grading as a calibration anchor and build a reusable LLM-as-judge system for large-scale evaluation.
Table~\ref{tab:human-graders} summarizes the experts recruited by our team for this human ground-truth calibration.

\begin{figure}[tbp]
\centering
\setlength{\abovecaptionskip}{0.1cm}
\includegraphics[width=\linewidth]{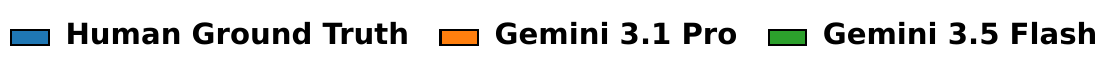}
\includegraphics[width=0.93\linewidth]{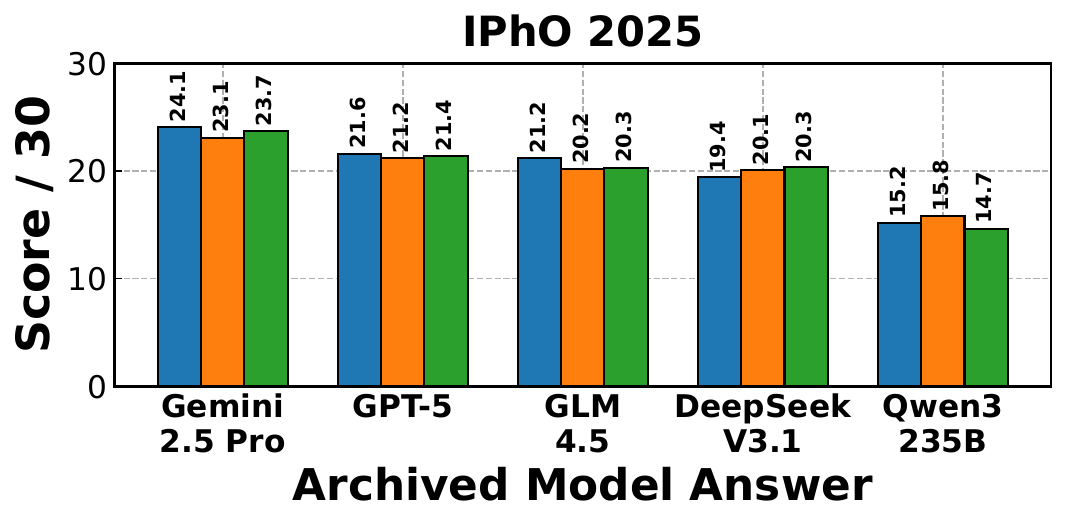}
\includegraphics[width=0.93\linewidth]{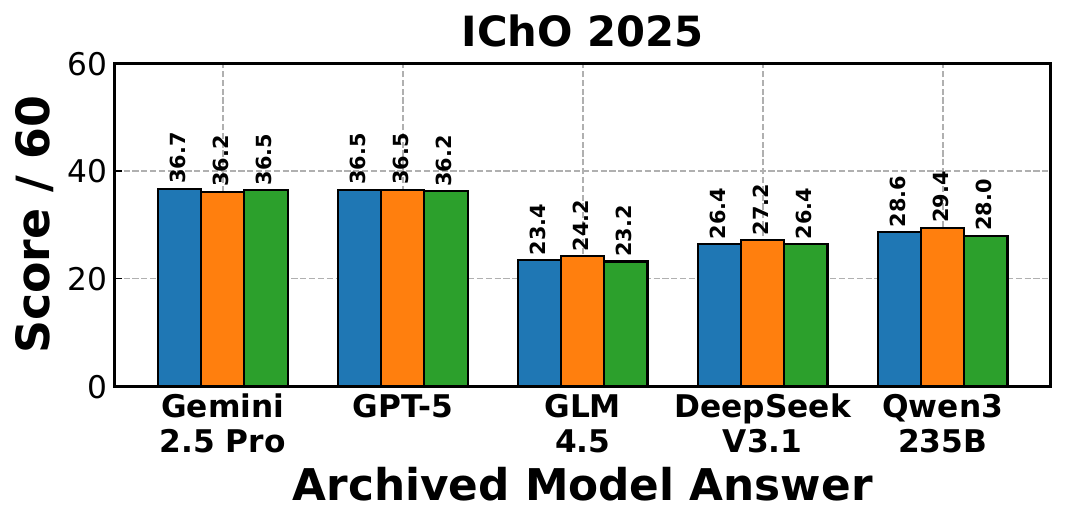}
\caption{Calibration of the LLM-as-judge system against human expert ground truth. The blue bar is the original score assigned by the human experts in Table~\ref{tab:human-graders}; the other bars show the two judge models.}
\label{fig:judge-calibration}
\vspace{-0.3cm}
\end{figure}

The judge is a pipeline, not a single prompt.
For each non-objective turn, it receives the staticized question, official answer, detailed solution, rubric, maximum score, and candidate response; here \emph{staticized} denotes a frozen textual rendering of the question-side problem content, including any visual evidence needed by a text-only route.
It acts as a strict olympiad examiner, assigns partial credit, and returns JSON with \texttt{total\_score}, \texttt{max\_score}, \texttt{sub\_scores}, evidence, deductions, and confidence.
Post-processing validates JSON, repairs common LaTeX escape errors, clamps impossible scores, checks maximum scores, and aggregates turns; objective-key items such as IBO are scored deterministically.
Each score remains traceable to a turn packet, rubric item, and evidence field, so expert-identified blind spots can be fixed locally.

For the four 2026 competition columns in Table~\ref{tab:latest-all-benchmarks}, we use a stricter sealed implementation with a fixed Gemini 3.5 Flash judge route. The judge receives the candidate response, provenance-labeled official answer and solution, the exact rubric criterion identifiers and maxima, and the original question/solution page PNGs directly when the frozen item supplies them. It must return bounded decimal sub-scores, evidence quotations, and an arithmetic-consistent total. Validation rejects rather than clamps an invalid response and stores one initial call plus at most six retries in a contiguous ledger; an exact solver-no-answer sentinel is deterministically scored zero without a judge call.

We calibrate the system on archived, non-regenerated answers from five models for IPhO 2025 and IChO 2025.
Gemini 3.5 Flash and Gemini 3.1 Pro both stay within one point of the medalist total score for every model--exam pair (Figure~\ref{fig:judge-calibration}).
Item-level replay further shows strong correlations with expert scores, indicating that the close total-score agreement is not driven solely by cancellation across items.
The final judge reached this agreement only after explicitly packaging expert rationales, requiring structured sub-scores, and preserving post-processing rules for known edge cases such as surface-similar chemistry answers and tiny subpart allocations. The resulting judge is a scalable proxy for medalist grading only when the official solution and rubric are available and correctly digitized.
It shifts the expensive human work from repeatedly grading every model to validating benchmark items and calibrating the judge.

\begin{figure}[t]
\setlength{\abovecaptionskip}{0.2cm}
\centering
\begin{tikzpicture}[
  font=\footnotesize,
  >={Latex[length=2mm]},
  node distance=3.5mm and 10mm,
  box/.style={draw, rounded corners, align=center, inner sep=2.5pt, minimum width=28mm},
  line/.style={-Latex, thick}
]

\node[box] (A0) {All subquestions\\$Q_{1..m}$};
\node[box, below=of A0] (A1) {LLM solves\\in one go};
\node[box, below=of A1] (A2) {Judge / Score};
\draw[line] (A0) -- (A1);
\draw[line] (A1) -- (A2);
\node[above=2mm of A0, font=\footnotesize\bfseries] {One-shot};

\node[box, right=12mm of A0.north east, anchor=north west] (B0) {Subquestion $Q_t$};
\node[box, below=of B0] (B1) {Context: $Q_{1..t}$\\+ prior answers $A_{1..t-1}$};
\node[box, below=of B1] (B2) {LLM outputs $A_t$};
\draw[line] (B0) -- (B1);
\draw[line] (B1) -- (B2);

\node[box, below=of B2] (B3) {Repeat for $t=1..m$};
\node[box, below=of B3] (B4) {Judge / Score};
\draw[line] (B2) -- (B3);
\draw[line] (B3) -- (B4);
\draw[line] (B3.east) -- ++(6mm,0) coordinate (loop)
            -- (loop |- B1.east) -- (B1.east);

\node[above=2mm of B0, font=\footnotesize\bfseries] {Interleaved};

\end{tikzpicture}
\caption{One-shot vs. interleaved solving schematic.}
\label{fig:one-shot-vs-iterative}
\vspace{-0.2cm}
\end{figure}
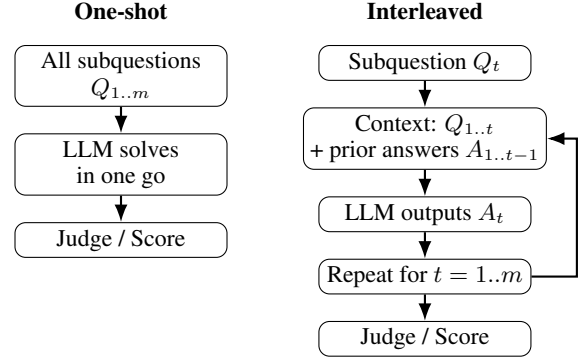

\begin{table}[t]
\vspace{-0cm}
\centering
\scriptsize
\setlength{\tabcolsep}{5pt}
\renewcommand{\arraystretch}{1.08}
\begin{tabular}{l|l|r r r r}
\toprule
\textbf{Model} & \textbf{Protocol} & \textbf{IPhO} & \textbf{IChO} & \textbf{CChO} & \textbf{CPhO} \\
\multicolumn{2}{l}{\textit{Full Scores}} & \textit{30.0} & \textit{60.0} & \textit{100.0} & \textit{320.0} \\
\midrule
\multirow{2}{*}{Gemini 3.5 Flash}
  & One-shot & 22.84 & 45.52 & 52.87 & 234.1 \\
  & Interleaved & \textbf{25.67} & \textbf{48.43} & \textbf{57.12} & \textbf{271.6} \\
\midrule
\multirow{2}{*}{Gemini 3.1 Pro}
  & One-shot & 24.19 & 42.02 & 50.94 & 242.4 \\
  & Interleaved & \textbf{27.30} & \textbf{46.45} & \textbf{57.44} & \textbf{302.1} \\
\midrule
\multirow{2}{*}{GPT-5.5}
  & One-shot & 20.80 & 33.52 & 39.48 & 285.6 \\
  & Interleaved & \textbf{25.63} & \textbf{33.98} & \textbf{46.59} & \textbf{289.5} \\
\midrule
\multirow{2}{*}{Claude Opus 4.7}
  & One-shot & 19.90 & 31.15 & 36.45 & 251.0 \\
  & Interleaved & \textbf{22.73} & \textbf{41.68} & \textbf{49.61} & \textbf{273.0} \\
\bottomrule
\end{tabular}
\vspace{-0.4em}
\caption{One-shot versus interleaved solving on open-ended runs across four models and four contest. Bold marks the higher score for each model--exam pair. Interleaved prompting wins all 16 displayed comparisons.}
\label{tab:oneshot-interleaved-ablation}
\vspace{-0.5cm}
\end{table}

\subsection{Solving Protocols}
\label{sec:solving-protocols}
Physics and chemistry exams often contain dependent subparts, so we compare two prompting protocols.
In \textbf{one-shot} solving, the model receives the full problem and writes one complete solution.
In \textbf{interleaved} solving, it answers one subpart at a time while previous turns remain in context.
Interleaving is not a new algorithm; it is a natural chat interaction pattern whose effect we test directly.

Figure~\ref{fig:one-shot-vs-iterative} illustrates the two protocols, and Table~\ref{tab:oneshot-interleaved-ablation} reports the archived comparison.
Interleaving wins all 16 displayed model--exam comparisons.
We use interleaved prompting for main open-ended evaluations; IBO is excluded because it is objective-key scored. If a solver response is empty, incomplete, or non-concrete—for example, because the reasoning-token limit is reached—the request is retried up to five times, and the first valid response is retained. If all five retries fail, the turn is marked as unsuccessful and this marker remains in the context for subsequent subparts. No correction is provided, and error-forward credit is awarded only when the response contains a concrete intermediate result.

\section{Experimental Results and Diagnostics}
\label{sec:3}

\subsection{Main Evaluation Setting}
We evaluate fourteen recent models from major model families on the thirteen-benchmark \textsc{ScienceArena} suite.
The model set includes the Gemini 3/3.5 family, GPT-5.4/5.5, Claude Opus 4.7, Doubao Seed 2.0 Pro, Qwen3.6/3.7, DeepSeek V4, MiniMax M2.7, GLM-5.1, and Xiaomi MiMo.
Model release dates are taken from official announcements when available---e.g., Gemini 3 Flash, Gemini 3.1 Pro, Gemini 3.5 Flash, GPT-5.4/5.5, Claude Opus 4.7, and Qwen3.7-Max \citep{google_gemini3flash_release_2025,google_gemini31_release_2026,google_gemini35_release_2026,openai_gpt54_release_2026,openai_gpt55_release_2026,anthropic_opus47_release_2026,qwen37_release_2026}---and otherwise from public availability reports or the provider registry snapshot used by our evaluation scripts.

Nine benchmark columns use a uniform \textbf{text-only interleaved} protocol.
For the four 2026 competition columns in Table~\ref{tab:latest-all-benchmarks}, input routing follows the verified capability of each exact API route: eight vision-capable routes receive native question text together with the exact original question-page PNG bytes directly in the solver API call, never an image description generated by another model; six genuine text-only routes receive an answer-blind question-only staticization and zero image blocks.
The text-staticization candidates are generated from question text and question-page images by Gemini 3.1 Pro, then reviewed item by item and corrected when needed before freezing; they contain no official answer, solution, or rubric. Because input mode is necessarily coupled to route capability, the four-column block is a capability evaluation rather than a controlled modality ablation.
Open-ended physics and chemistry turns are graded by the calibrated LLM-as-judge system from Section~\ref{sec:judge-system}.
IBO is the exception: it is objective-key scored and is evaluated with the dedicated IBO runner over the original keyed items rather than by an LLM judge.

For international competitions, we annotate model scores with human reference medals.
The references are theory-score aggregates from the official rankings: IPhO 2025 winner/gold/silver/bronze are 29.2/23.4/17.0/11.5 out of 30; IChO 2025 are 57.1/36.6/28.7/26.3 out of 60; IBO 2023 are 395.4/354.7/293.1/249.7 out of 453.
We do not assign medals to national competitions, because national selection pools and award rules are not directly comparable across countries.

\begin{table}[t]
\centering
\small
\setlength{\tabcolsep}{5pt}
\begin{tabular}{lrrr}
\toprule
\textbf{Metric} & \textbf{IPhO 25} & \textbf{IChO 25} & \textbf{IBO 23} \\
\textit{Full score} & \textit{30.0} & \textit{60.0} & \textit{453.0} \\
\midrule
\multicolumn{4}{l}{\textit{\textbf{Human References}}} \\
\quad \Winner{Human No.1} & 29.2 & 57.1 & 395.4 \\
\quad \Gold{Human Gold}  & 23.4 & 36.6 & 354.7 \\
\quad \Silver{Human Silver} & 17.0 & 28.7 & 293.1 \\
\quad \Bronze{Human Bronze} & 11.5 & 26.3 & 249.7 \\
\midrule
\multicolumn{4}{l}{\textit{\textbf{Archived Five-Model Performance}}} \\
\quad Gemini 2.5 Pro
  & \Gold{\best{{24.1}}}
  & \Gold{\best{{36.7}}}
  & \Winner{\best{{404.0}}} \\
\quad GPT-5
  & \Silver{21.8}
  & \Silver{31.8}
  & \Gold{{395.0}} \\
\quad GLM 4.5
  & \Silver{21.2}
  & 19.8
  & \Silver{329.0} \\
\quad DeepSeek V3.1
  & \Silver{19.4}
  & \Bronze{26.6}
  & \Silver{312.0} \\
\quad Qwen3 235B
  & \Bronze{15.2}
  & 25.9
  & \Silver{337.0} \\
\bottomrule
\end{tabular}
\vspace{-0.1cm}
\caption{Human reference scores used for medal annotations on international competitions. IPhO and IChO use theory-only scores; IBO uses the objective-key total used by our evaluation runner. All five models were released before the start of the competition.}
\label{tab:main_results}
\vspace{-0.5cm}
\end{table}

\begin{table*}[t]
\centering
\scriptsize
\setlength{\tabcolsep}{2.1pt}
\renewcommand{\arraystretch}{1.08}
\resizebox{\textwidth}{!}{%
\begin{tabular}{rrrrrrrrrrrrrrl}
\toprule
\makecell{\textbf{Contest}\\\textbf{Release}\\\textbf{Max. Score}} & \makecell{IBO\\25-08\\453} & \makecell{APhO\\25-05\\30} & \makecell{EuPhO\\25-06\\30} & \makecell{USNCO\\25-06\\100} & \makecell{IPhO\\25-07\\30} & \makecell{IChO\\25-07\\60} & \makecell{CPhO\\25-10\\320} & \makecell{CChO\\25-10\\100} & \makecell{INChO\\26-01\\106} & \makecell{NBPhO\\26-04\\72} & \makecell{USAPhO\\26-05\\150} & \makecell{IPhO\\26-07\\30} & \makecell{IChO\\26-07\\60} & \makecell{Weighted\\Avg.\\(Rank)} \\
\midrule
\arrayrulecolor{red!75!black}\cline{2-9}\arrayrulecolor{black}
\textbf{Gemini 3 Flash} (25-12) & \Winner{428.00} & 26.75 & \RankOne{23.60} & 93.50 & \Gold{25.20} & \Gold{44.74} & \RankThree{300.50} & \RankThree{59.55} & \ContamV{81.50} & 46.80 & \RankOne{150.00} & 28.00 & \Gold{50.35} & \RankThree{83.6 (\#3)} \\
\arrayrulecolor{red!75!black}\cline{10-10}\arrayrulecolor{black}
\rowcolor{blue!6}\textbf{Seed 2.0 Pro} (26-02) & \Winner{415.00} & 21.72 & 13.60 & 72.30 & \Silver{18.92} & \Gold{37.14} & 243.70 & 53.18 & 87.20 & \ContamV{35.60} & 147.00 & 20.75 & \Silver{40.50} & 69.4 (\#11) \\
\textbf{Gemini 3.1 Pro} (26-02) & \Winner{431.00} & \RankOne{28.50} & \RankTwo{22.50} & \RankOne{97.00} & \Winner{29.45} & \Gold{48.95} & \RankTwo{303.50} & \RankOne{72.28} & \RankOne{95.95} & \ContamV{48.30} & 146.00 & \RankOne{29.75} & \Gold{54.18} & \RankOne{88.7 (\#1)} \\
\rowcolor{blue!6}\textbf{GPT-5.4} (26-03) & \Gold{358.00} & 22.01 & 14.30 & 76.90 & \Silver{20.94} & \Silver{31.70} & 235.30 & 47.79 & 83.50 & \ContamV{37.60} & 145.00 & 21.95 & \Silver{31.87} & 67.3 (\#12) \\
\textbf{GLM-5.1} (26-04) & \Silver{349.00} & \RankTwo{28.40} & 8.20 & 90.00 & \Gold{26.11} & \Gold{38.82} & 283.00 & 58.61 & 85.75 & \ContamV{48.60} & \RankOne{150.00} & \RankOne{29.75} & \Gold{48.54} & 78.2 (\#6) \\
\rowcolor{blue!6}\textbf{Qwen3.6-Plus} (26-04) & \Winner{413.00} & 27.56 & 13.30 & 88.20 & \Silver{22.87} & \Silver{36.22} & 252.80 & 49.33 & 79.25 & \ContamV{46.80} & 149.25 & 27.25 & \Gold{45.12} & 75.8 (\#8) \\
\textbf{MiniMax M2.7} (26-04) & \Silver{337.00} & 22.56 & 8.20 & 71.80 & \Silver{19.64} & 26.06 & 184.70 & 43.78 & 82.75 & \ContamV{35.70} & 139.50 & 27.00 & \Bronze{27.51} & 62.7 (\#14) \\
\rowcolor{blue!6}\textbf{Claude Opus 4.7} (26-04) & \Winner{407.00} & 24.19 & 16.50 & 75.20 & \Silver{22.29} & \Gold{44.49} & 262.00 & 56.59 & 82.70 & \ContamV{48.70} & 132.50 & 26.05 & \Gold{45.99} & 75.8 (\#9) \\
\textbf{MiMo V2.5 Pro} (26-04) & \Silver{317.00} & 21.33 & 11.70 & 74.00 & \Silver{18.97} & 28.94 & 248.10 & 50.11 & 86.00 & \ContamV{26.30} & 139.50 & 22.80 & \Silver{29.83} & 63.8 (\#13) \\
\rowcolor{blue!6}\textbf{GPT-5.5} (26-04) & \Winner{415.00} & 21.50 & 19.00 & 75.40 & \Silver{22.08} & \Silver{35.77} & 270.30 & 53.30 & 83.00 & \ContamV{\RankOne{59.20}} & \RankOne{150.00} & \RankOne{29.75} & \Winner{55.12} & 78.8 (\#5) \\
\textbf{DeepSeekV4-Pro} (26-04) & \Silver{327.00} & 26.01 & 13.30 & 90.00 & \Gold{25.10} & \Silver{35.82} & 257.70 & 49.78 & \RankThree{89.75} & \ContamV{\RankThree{51.40}} & 148.25 & \RankOne{29.75} & \Gold{48.98} & 77.1 (\#7) \\
\rowcolor{blue!6}\textbf{DeepSeekV4-Flash} (26-04) & \Silver{324.00} & 20.44 & \RankThree{20.40} & \RankThree{94.50} & \Silver{20.53} & \Gold{38.33} & 182.90 & 46.46 & 83.80 & \ContamV{\RankTwo{51.50}} & \RankOne{150.00} & 27.75 & \Gold{49.46} & 74.1 (\#10) \\
\arrayrulecolor{red!75!black}\cline{11-12}\arrayrulecolor{black}
\textbf{Gemini3.5-Flash} (26-05) & \Winner{430.00} & 27.91 & 16.80 & \RankTwo{95.70} & \Gold{26.41} & \Gold{47.98} & \RankOne{307.00} & \RankTwo{70.60} & \RankTwo{92.25} & 49.30 & 148.50 & \ContamV{\RankOne{29.75}} & \Winner{54.91} & \RankTwo{86.1 (\#2)} \\
\rowcolor{blue!6}\textbf{Qwen3.7-Max} (26-05) & \Silver{349.00} & \RankThree{28.10} & 15.60 & 93.00 & \Gold{26.52} & \Gold{41.28} & 277.50 & 52.66 & 83.00 & 46.80 & 148.00 & \ContamV{\RankOne{29.75}} & \Gold{46.68} & 79.3 (\#4) \\
\arrayrulecolor{red!75!black}\cline{13-14}\arrayrulecolor{black}
\bottomrule
\end{tabular}%
}
\vspace{-0.2cm}
\caption{Main evaluation on thirteen olympiad-style benchmarks. Dates use YY-MM, and rows use provider family/API dates. The \strong{red stair-step} marks a descriptive provider-date boundary, including the boundary between the May 2026 model rows and the July 2026 IPhO/IChO columns; it is not proof of absent training exposure because exact evaluated-route availability and training cutoffs are unavailable. The four 2026 columns are therefore reported as public-benchmark capability measurements. Medal icons use the international reference cutoffs in Table~\ref{tab:main_results} and are shown only for IPhO 2025, IChO 2025\&26, and IBO. For other competitions, \RankOne{gold}, \RankTwo{blue}, and \RankThree{green} mark \RankOne{first}, \RankTwo{second}, and \RankThree{third} place within the column. Weighted average normalizes each score to 0--100 and averages them.}
\label{tab:latest-all-benchmarks}
\vspace{-0.4cm}
\end{table*}

\subsection{Overall Trends}
Table~\ref{tab:latest-all-benchmarks} shows that frontier models now reach extremely high scores on several olympiad-style exams, but the picture is not uniform.
Table~\ref{tab:main_results} provides the human reference points used to interpret the three international columns in the main table.
On IPhO 2025, Gemini 3.1 Pro reaches \Winner{29.45}/30, slightly above the best human theory score in the official ranking.
Several other models, including Gemini 3.5 Flash, Gemini 3 Flash, Qwen3.7-Max, GLM-5.1, and DeepSeek V4 Pro, clear the IPhO gold reference.
On IChO 2025, the best current score is Gemini 3.1 Pro at \Gold{48.95}/60: well above the gold threshold but still below the human winner score of 57.1.
On IBO 2023, the current best models exceed the human-winner reference, with the Gemini family occupying the top three scores in our objective-key run.
Thus, the strongest current models obtain gold-medal-equivalent rubric scores across physics, chemistry, and biology on these public benchmarks, but exceeding the human-winner reference is domain-dependent: it occurs on IPhO and IBO in this suite, while the hardest IChO structures still leave a substantial gap to the human winner.

Across the full suite, Gemini models dominate the most stable comparisons.
Using equal-weighted benchmark averages over models with complete coverage, Gemini 3.1 Pro is strongest in physics (89.5\% of full score), chemistry (86.3\%), and biology (431.00/453, 95.1\%).
The result is similar under point-weighted aggregation, where Gemini 3.1 Pro also remains first in physics after the 2026 results are included.
The three Gemini variants take every top-three position on IChO 2025 and IBO, and two of the three top positions on IPhO 2025.
OpenAI and Anthropic models are competitive on several individual exams--GPT-5.5 leads across the four 2026 columns in Table~\ref{tab:latest-all-benchmarks} and Claude Opus 4.7 remains strong on IChO 2025--but neither matches the Gemini 3.1/3.5 pair across the full suite.
The strongest non-Gemini outliers are informative: Qwen3.7-Max is second on IPhO, GLM-5.1 is near the top on APhO/USAPhO, and DeepSeek V4 Flash is surprisingly strong on USNCO despite weaker results on physics tracks.

The red release screen is essential for interpreting these numbers, but it is deliberately descriptive.
It compares documented provider family/API dates with public exam-material months; it does not establish the first availability date or backend revision of the exact evaluated 360 route, and pre-release timing alone would not prove absence from training data.
Post-release and temporally unknown cells therefore measure \emph{current capability under a public benchmark}, not a strict time-release holdout.
The only narrow release-based holdout we identify is Gemini 2.5 Pro: its stable 17 June 2025 release predates the IChO (10 July) and IPhO (21 July) theory examinations, and expert grading yields 36.71/60 and 24.1/30, just above the respective gold references of 36.6 and 23.4 \citep{google_gemini25_stable_2025,icho2025_official_program,ipho2025_official_schedule}.
We do not extend this observation to IBO or treat release timing as proof of absent training exposure.
All 182 model--benchmark cells in Table~\ref{tab:latest-all-benchmarks} are populated; its four 2026 columns contribute 56 complete cells, with no imputation for route, generation, or judge failures.

The sealed four-contest block comprises NBPhO, USAPhO, IPhO, and IChO 2026 under the protocol described above.
GPT-5.5 has the highest four-contest normalized macro score at 93.31\%, followed by Gemini 3.5 Flash at 89.54\% and Gemini 3.1 Pro at 88.47\%.
This four-contest ranking is not the same estimand as the thirteen-column average: it covers only those four sealed 2026 contests, whereas the final column of Table~\ref{tab:latest-all-benchmarks} also includes nine text-only benchmark measurements.
Appendix Table~\ref{tab:fresh-2026-ci} reports every route's four-contest macro score, sequence-cluster interval, and input assignment. Readers can use its V/T column to identify whether an exact model route was evaluated with native images (V) or as text-only (T).

\stepcounter{table}
\subsection{Solver Response Stability Across Repeated API Calls}
We repeat the complete solver-and-scoring call three times on IPhO and IChO 2026 while holding the exact API route, input packet, prompt, rubric, and fixed Gemini 3.5 Flash judge configuration constant. These recent, demanding open-ended exams provide a direct stress test of response stability under ordinary API use; the frozen-answer cross-judge experiment below isolates judge variation.

\begin{table}[t]
\centering
\scriptsize
\setlength{\tabcolsep}{4.2pt}
\begin{tabular}{l r r r}
\toprule
\textbf{Contest metric} & \textbf{Gemini 3.5 Flash} & \textbf{GPT-5.5} & \textbf{Claude Opus 4.7} \\
\midrule
IPhO mean (/30) & 28.75 & 29.75 & 26.75 \\
IPhO SD & 1.00 & 0.00 & 1.32 \\
IPhO range & 2.00 & 0.00 & 2.50 \\
\midrule
IChO mean (/60) & 53.73 & 54.84 & 46.45 \\
IChO SD & 0.92 & 0.80 & 2.64 \\
IChO range & 1.77 & 1.50 & 4.95 \\
\bottomrule
\end{tabular}
\vspace{-0.2cm}
\caption{Contest-level stability over five independent end-to-end calls per exact model route. Each call regenerates all answers and re-runs the fixed Gemini 3.5 Flash judge; SD is the sample standard deviation.}
\label{tab:fresh-2026-run-stability}
\vspace{-0.4cm}
\end{table}

The three routes remain in the same high-performance regime across repeated calls. GPT-5.5 is especially stable (zero IPhO spread and a 1.50/60 IChO range); Gemini 3.5 Flash spans 2.00/30 and 1.77/60. Claude Opus 4.7 has the largest variation, yet its maximum span is 2.50/30 on IPhO and 4.95/60 on IChO. Thus the API-level contest scores are operationally repeatable, although stability is provider- and exam-dependent. In a separate frozen-answer experiment, exact three-repeat item agreement across six judges ranges from 90.11\% to 96.70\%; Appendix Table~\ref{tab:cross-family-judge-stability} reports the per-judge totals and isolates judge-family sensitivity.

\subsection{Controlled Modality Ablation}
To isolate input modality from route capability, we evaluate 12 verified vision-capable routes on 12 image-dependent subquestions (four each from IPhO 2025, IChO 2025, and IBO 2023) under native-image, faithful audited-description, and no-visual-information conditions, with four repeats per arm.
Open-ended physics and chemistry cells use the mean of three rubric judges, whereas IBO are scored deterministically against the objective key.

\begin{table}[t]
\centering
\small
\setlength{\tabcolsep}{5pt}
\begin{tabular}{l r}
\toprule
\textbf{Condition} & \textbf{Mean normalized score} \\
\midrule
Native image & 72.75\% \\
Faithful textual description & 71.73\% \\
No visual information & 49.85\% \\
\bottomrule
\end{tabular}
\vspace{-0.2cm}
\caption{Controlled modality ablation over contests.}
\label{tab:controlled-modality-ablation}
\vspace{-0.5cm}
\end{table}

Native images outperform no visual information by 22.90 percentage points (paired-bootstrap 95\% CI [20.02, 25.77]), showing that visual evidence materially affects performance on this deliberately image-dependent panel.
The overall native-image--description estimate is 1.02 points and statistically unresolved (95\% CI [$-1.20$, 3.27]); chemistry nevertheless shows a 5.01-point native-image advantage (95\% CI [0.57, 9.56]).
These results also support a practical composition for text-only solvers: a multimodal captioner can first convert the image into a faithful, quality-controlled description, after which the text model recovers most of the native-image performance. This captioner--solver pipeline therefore provides strong multimodal problem-solving capability when direct vision input is unavailable, while native pixels remain preferable for structure-sensitive chemistry.
Domain-specific estimates are reported in Appendix Table~\ref{tab:modality-domain-split}.

\subsection{Subfield and Modality Diagnostics}
\begin{figure*}[t]
\centering
\includegraphics[width=0.95\textwidth]{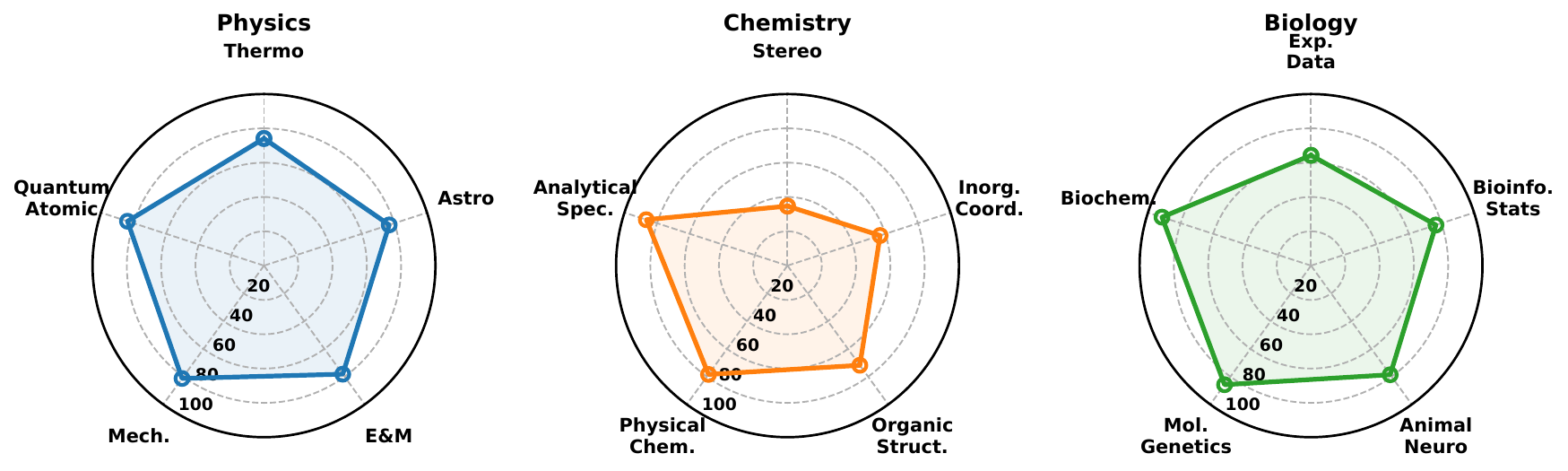}
\vspace{-0.6em}
\caption{Subfield diagnostics aggregated over clean model--benchmark cells. Each radar uses five domain-specific axes, so polygon shapes should be read within a domain rather than across domains. Values are awarded points divided by available points. Together, the plots localize which kinds of scientific reasoning remain most fragile.}
\label{fig:subfield-diagnostics}
\vspace{-0.35cm}
\end{figure*}

Aggregate scores hide where models fail.
We therefore classify each benchmark turn into a scientific subfield and skill bucket using a Gemini 3.5 Flash classification pass over the staticized question, official answer, and rubric.
The classifier never sees model answers and does not change the benchmark.
We then join these labels with clean turn-level grades to measure score rates by subfield, image dependence, structure dependence, and quantitative reasoning.
Figure~\ref{fig:subfield-diagnostics} summarizes five representative subfields for each domain; the full numeric table, including coverage counts, is in Appendix~\ref{app:subfield_table}.
Across domains, representative failures share fragile binding between problem evidence, the scoring target, and the required answer form; Figure~\ref{fig:structure-failure-cases} gives one case per domain, with more examples in Appendix~\ref{app:additional_cases}.
These coverage counts are model--turn grading instances rather than independent questions; slices with small counts are exploratory.

\begin{figure*}[t]
\centering
\setlength{\abovecaptionskip}{0cm}
\begin{tcolorbox}[
  title={Case A: Global physical invariant missed after fluent local reasoning (IPhO 2025 S2 C1--C2)},
  colback=gray!3,
  colframe=black!35,
  boxrule=0.6pt,
  arc=1.5mm,
  left=1.5mm,right=1.5mm,top=1mm,bottom=1mm
]
\footnotesize
\textbf{Task.} Reason about a coupled mechanical system described over multiple subparts and identify the correct force/energy regime.\\
\textbf{Model behavior.} GPT-5 gave detailed equations, but the human expert marked C1 and C2 as 0/1 because the answer missed the conserved total-force constraint and did not establish the correct physical picture.\\
\textbf{Diagnostic.} The failure is not algebraic fluency; it is global model control. Once the response anchors on the wrong regime, later derivations become locally coherent but physically irrelevant.
\end{tcolorbox}

\vspace{-0.5em}

\begin{tcolorbox}[
  title={Case B: Stereochemical template applied to the wrong substrate (IChO 2025 Q2.2a--b)},
  colback=gray!3,
  colframe=black!35,
  boxrule=0.6pt,
  arc=1.5mm,
  left=1.5mm,right=1.5mm,top=1mm,bottom=1mm
]
\footnotesize
\textbf{Task.} Use a Zimmerman--Traxler transition-state model to complete an Evans aldol step and draw the major product.\\
\textbf{Model behavior.} Gemini 3.5 Flash produced a chemically fluent aldol explanation, but treated the enolate as a generic methyl-ketone/propionate case and placed substituents in the wrong positions.\\
\textbf{Score.} The judge awarded 0/15 for the transition-state placement and 0/4 for the product in the affected subpart.\\
\textbf{Diagnostic.} The model knows the named reaction pattern, but it fails at exact structure binding: reading the actual reactants, preserving the auxiliary, and propagating stereochemistry into a drawable product.
\end{tcolorbox}

\vspace{-0.5em}

\begin{tcolorbox}[
  title={Case C: Biology knowledge is high, but multi-statement data consistency fails (IBO 2023 Item 63)},
  colback=gray!3,
  colframe=black!35,
  boxrule=0.6pt,
  arc=1.5mm,
  left=1.5mm,right=1.5mm,top=1mm,bottom=1mm
]
\footnotesize
\textbf{Task.} Interpret nutrient-depletion experiments on \textit{Symbiodinium} proliferation and differentiation and answer five true/false statements.\\
\textbf{Model behavior.} Strong models commonly returned partial sequences such as F F F F T or F T F F F instead of the official F T T F T. They often noticed proliferation changes but misread which nutrient manipulations changed stage ratios.\\
\textbf{Score.} Across six representative frontier models, the average score on this item was 2.83/5.\\
\textbf{Diagnostic.} IBO errors usually arise when several statements must be checked against the same experimental figure. The model may explain each statement plausibly, but one local interpretation breaks cross-panel consistency.
\end{tcolorbox}
\caption{Representative cross-domain failure cases. The common theme is not absence of scientific terminology, but fragile binding between the problem evidence, the scoring target, and the final answer format. These cases expose errors that answer-only accuracy would often obscure. Additional cases appear in Appendix~\ref{app:additional_cases}.}
\label{fig:structure-failure-cases}
\vspace{-0.5cm}
\end{figure*}

The strongest pattern is that \textbf{chemistry is bottlenecked by structure fidelity}.
Physical chemistry is comparatively strong (78.4\%), while stereochemistry is much lower (34.7\%).
Organic-structure questions are also substantially weaker (71.8\%) than analytical/spectroscopy questions (86.6\%).
More generally, chemistry turns that require structural diagrams score 64.5\%, compared with 74.1\% for chemistry turns without such requirements; visually grounded chemistry turns score 63.3\%, compared with 83.1\% for non-visual chemistry turns.
This matches our qualitative inspection: models often know the relevant reaction logic but lose credit when they must commit to an exact connectivity, stereochemical relation, spectrum-to-structure inference, or diagram-derived intermediate.
The error is rarely a simple lack of chemistry vocabulary.
Instead, models often retrieve a plausible named mechanism or biosynthetic template, then fail to bind that template to the substrate, substituent positions, and stereochemical constraints in problems.

Physics shows a different and more globally coupled failure profile. Thermodynamics and astrophysics are the weakest physics slices in Figure~\ref{fig:subfield-diagnostics}, while mechanics, electromagnetism, and modern/ atomic physics are all in the 78--84\% range. The visual/non-visual gap is small (77.6\% vs.\ 79.7\%). The main physics errors are usually global: a wrong early approximation or variable definition propagates through later subparts. Interleaving helps because each subpart receives focused attention, but it cannot fully repair a mistaken physical model once the conversation has anchored on it.

Biology is the highest-scoring international track.
Seven models exceed the IBO human-winner reference, and the best score is Gemini 3.1 Pro at 431.00/453.
This does not mean biology reasoning is solved: IBO is objectively keyed, which makes grading less fragile than process-rubric chemistry, and many questions reward careful elimination among constrained options.
Within IBO, models are strongest on biochemistry (91.4\%) and molecular genetics (85.8\%), but weaker on experimental-data interpretation (64.3\%) and bioinformatics/statistics (76.6\%).
The remaining errors are less about missing factual knowledge and more about multi-statement consistency, quantitative interpretation, and careful reading of experimental setups.

\subsection{Insights from Gold-Medalist Grading}
\label{sec:expert-insights}
The archived gold-medalist grading for five model submissions on IPhO/IChO 2025 separates scientifically fluent prose from contest-credit evidence.
IPhO notes expose fragile global reasoning after strong local derivations, whereas IChO notes require explicit carbon skeletons, stereochemistry, and valid error-forward structures; these observations motivate our rubric gates for answer form, structure specificity, and subpart evidence.
The provenance-preserving export contains 525 expert-scored rows---410 with a nonempty medalist comment and 115 explicitly marked as having no comment---and the longer note archive appears in Appendix~\ref{app:expert_notes}.
A deterministic replay of 1,050 stored rows from the two calibrated Gemini judges gives interjudge normalized MAE 0.093 and Pearson correlation 0.907 across 525 shared expert rows; this remains a narrow same-family diagnostic.

\vspace{-0.05cm}

\subsection{Judge Stability and Protocol Diagnostics}
The human calibration in Figure~\ref{fig:judge-calibration} shows close agreement between the judge pipeline and medalist scores. Structure-heavy chemistry is the most demanding case because ambiguous candidate prose may not uniquely specify the required molecule; when resources permit, expert adjudication provides the strongest supplementary safeguard for low-confidence or structure-only responses.

Exact three-repeat item agreement ranges from 90.11\% to 96.70\% across the six judges. Among the four direct-PNG vision judges, the six judge pairs have mean normalized MAE 1.732 percentage points, mean Pearson correlation 0.888, and mean Spearman correlation 0.843. The two text-staticized judges are reported separately as modality-sensitivity conditions. Together with the expert calibration, these results demonstrate a repeatable, cross-family-stable judging pipeline under matched inputs. Per-judge totals and strict-replay details are reported in Appendix Table~\ref{tab:cross-family-judge-stability}.

Table~\ref{tab:oneshot-interleaved-ablation} favors interleaved prompting in all 16 displayed model--exam comparisons across four scientific examinations, providing consistent end-to-end evidence for decomposing multi-part problems into focused turns while retaining relevant reasoning context throughout extended solutions. Interleaving approximates multi-turn use; tool, retrieval, and verifier-agent stacks also evaluate orchestration and therefore belong in a separate track.

\section{Conclusion}

\vspace{0.1cm}

\textsc{ScienceArena} evaluates frontier LLMs on olympiad-style science problems with multi-step reasoning, public source provenance, and rubric-faithful scoring.
The benchmark spans thirteen recent international, regional, and national competitions; preserves a common interleaved interaction structure while directly routing original images to vision-capable models in the four 2026 competition columns; calibrates LLM-as-judge against gold-medalist human grading; and supports a broad latest-model comparison with explicit release-time risk metadata.
The results show that recent models can obtain medal-equivalent rubric scores on public international open-ended exams, with Gemini 3.1 Pro reaching the IPhO 2025 winner reference under the text-only interleaved protocol.
On the IBO track, several recent models exceed the human-winner score, while IChO remains below the best human contestant despite strong gold-level scores.

The diagnostic picture is as important as the headline scores.
Physics errors concentrate in global regime control and graph/figure binding; chemistry errors concentrate in exact structure and stereochemistry; biology errors concentrate in multi-statement data consistency rather than factual recall.
Human expert comments confirm that scientific fluency is not enough: contest credit depends on explicit, checkable commitments that match the rubric.
Because many latest-model results are post-release relative to the public exams, we report them as transparent current-capability measurements rather than strict uncontaminated holdouts.
Time-locked competitions, molecule-aware judging, and richer agentic protocols constitute complementary evaluation tracks; the simple interleaved API protocol provides a reproducible baseline for comparing model reasoning ability.

\clearpage
\section*{Limitations}
The benchmark spans thirteen competitions, but results may not generalize to all scientific curricula.
We evaluate theory components only, since experimental sections require physical apparatus and proctoring.
Expert grading is costly and currently covers judge calibration on IPhO/IChO. Across those contests, expert agreement and the cross-family stability study support the automated pipeline; contest-specific expert spot checks remain the strongest standard for high-stakes comparisons on additional contests.
The three repeated calls jointly measure solver-generation and fixed-route judge variation rather than separating them. The frozen-answer cross-judge study isolates judge variation for one solver on two contests. Capability-conditioned vision/text route comparisons jointly reflect route and modality; Table~\ref{tab:controlled-modality-ablation} provides the controlled input-modality estimate.
Finally, our subfield taxonomy is model-assisted and may reflect classifier priors; alternative decompositions could yield different diagnostic slices.


\bibliography{custom}

\appendix

\section{Additional Failure Cases}
\label{app:additional_cases}
Figure~\ref{fig:structure-failure-cases} shows one representative failure from each domain.
Here we provide one additional case per domain to make the qualitative diagnosis more complete.

\begin{tcolorbox}[
  title={Physics: correct theory, wrong graph binding (IPhO 2025 S1 B2)},
  colback=gray!3,
  colframe=black!35,
  boxrule=0.6pt,
  arc=1.5mm,
  left=1.5mm,right=1.5mm,top=1mm,bottom=1mm
]
\footnotesize
The human expert noted that GPT-5's theoretical analysis was correct, but the model used the wrong point on the correct curve.
This earned 0.3/0.5 rather than full credit.
The case illustrates a recurring physics pattern: symbolic derivation can be near-perfect while the final numerical answer is damaged by one visual-grounding decision.
\end{tcolorbox}

\begin{tcolorbox}[
  title={Chemistry: plausible mechanism, wrong intermediates (CChO 2025 Q9.7)},
  colback=gray!3,
  colframe=black!35,
  boxrule=0.6pt,
  arc=1.5mm,
  left=1.5mm,right=1.5mm,top=1mm,bottom=1mm
]
\footnotesize
The task asks for key intermediates in a Trigonoliimine C rearrangement.
Representative Gemini runs described a plausible spirocyclopropane/spiroindolenium-style pathway, but did not provide the official intermediate structures and mismatched the rearrangement sequence, receiving 0/6.
The failure is retrieval-adjacent: the prose resembles a real mechanism, but the scored object is an exact set of structures.
\end{tcolorbox}

\begin{tcolorbox}[
  title={Biology: experimental-design matching error (IBO 2023 Item 20)},
  colback=gray!3,
  colframe=black!35,
  boxrule=0.6pt,
  arc=1.5mm,
  left=1.5mm,right=1.5mm,top=1mm,bottom=1mm
]
\footnotesize
The item asks models to match four guppy mate-choice hypotheses to four experimental designs.
Several strong models returned 2 4 3 1 instead of the official 2 4 1 3, confusing the control for male activity/preference with the control for general side or group preference.
The answer is not factually obscure; the difficulty is maintaining the causal role of each experimental manipulation across four similar alternatives.
\end{tcolorbox}

\section{Full Subfield Diagnostics}
\label{app:subfield_table}
Table~\ref{tab:subfield-diagnostics-full} reports the numeric values underlying Figure~\ref{fig:subfield-diagnostics} in Section~\ref{sec:3}.
The main-text radar makes the within-domain comparison visually explicit; its values and coverage counts are tabulated below.

\begin{table}[h]
\centering
\small
\setlength{\tabcolsep}{3.5pt}
\begin{tabular}{llrr}
\toprule
\textbf{Domain} & \textbf{Subfield} & \textbf{Score} & \textbf{N} \\
\midrule
Phys. & thermodynamics & 74.1\% & 805 \\
Phys. & astrophysics & 76.8\% & 470 \\
Phys. & electromagnetism & 78.2\% & 258 \\
Phys. & mechanics & 81.3\% & 354 \\
Phys. & modern quantum atomic & 83.7\% & 332 \\
Chem. & stereochemistry & 34.7\% & 187 \\
Chem. & inorganic coordination & 56.8\% & 383 \\
Chem. & organic structure & 71.8\% & 419 \\
Chem. & physical chemistry & 78.4\% & 719 \\
Chem. & analytical spectroscopy & 86.6\% & 60 \\
Bio. & experimental data & 64.3\% & 14 \\
Bio. & bioinformatics statistics & 76.6\% & 42 \\
Bio. & animal behavior neurobiology & 78.6\% & 56 \\
Bio. & molecular genetics & 85.8\% & 266 \\
Bio. & biochemistry & 91.4\% & 70 \\
\bottomrule
\end{tabular}
\caption{Numeric subfield diagnostics corresponding to Figure~\ref{fig:subfield-diagnostics}. Score is total awarded points divided by total available points; $N$ counts model--turn grading instances, not independent questions. Small strata, especially biology experimental data ($N=14$), are exploratory.}
\label{tab:subfield-diagnostics-full}
\end{table}

The available diagnostic table records model--turn counts but lacks a question-level classification map; unique-question counts are therefore unavailable and are not inferred from $N$.

\section{Human Expert Notes and Ability Boundary}
\label{app:expert_notes}
This appendix expands the shortened discussion in Section~\ref{sec:expert-insights}.
The notes come from archived expert grading of the original five model answers on IPhO 2025 and IChO 2025.
They are useful because the graders comment on why a response deserves or loses partial credit, not merely on whether the final answer is correct.
We therefore use the same notes to define the ability-boundary radar plots in this appendix: symbolic derivation, graph/visual grounding, global coherence, structure/stereochemistry, analytical reasoning, and data interpretation are not abstract categories, but recurring distinctions made by the human graders.

\paragraph{Fluent derivation is not the same as physical control.}
In IPhO 2025, GPT-5 often received full credit for local symbolic derivations, and experts explicitly described several early subparts as essentially perfect.
The same solution then lost credit on graph- or regime-dependent items.
For example, in S1 B2 the expert noted that the theoretical analysis was correct, but the model used the wrong point on the correct curve, leading to only 0.3/0.5.
In S1 C4, the formula work was again strong, but the peak-frequency reading and statistical inference were off, giving 0.3/0.6.
The deeper lesson is that physics olympiad grading is not just equation checking: the model must bind a symbol to the correct plotted quantity, choose the right curve, and propagate measured values with the intended uncertainty.

\paragraph{Long reasoning can hide a wrong global model.}
The clearest IPhO failure appears in S2 C1--C2.
The human grader marked GPT-5 at 0/1 and 0/1 because the answer failed to realize a conserved total-force constraint and did not provide the necessary case analysis.
This is a different failure from arithmetic error.
The model can write a long solution, but if the first physical invariant is wrong, the response becomes a plausible derivation of the wrong problem.
This observation motivates our preference for interleaved evaluation: it gives each subpart focused context, but it also shows that interleaving alone cannot fix a mistaken regime choice once the conversation has anchored on it.

\paragraph{Chemistry experts grade exact structures, not chemical vibes.}
The IChO comments are even sharper.
On IChO 2025 Q1.3, Gemini 2.5 Pro suggested that the molecular formula might contain a typo and then used the wrong carbon skeleton; the expert marked the subpart 0 and identified this as a structure-comprehension failure rather than a minor wording issue.
On Q1.6, the same model described reaction logic that was partly reasonable, but repeatedly used a decalin-like skeleton instead of the required tricyclic skeleton; without a valid error-forward path, I/J/K all received 0.
For GPT-5, the expert comments similarly flagged failures to recognize a pictured compound, wrong methyl positions in a carbon skeleton, and incorrect nucleophile approach geometry.
These cases explain why chemistry remains harder than physics in Figure~\ref{fig:subfield-diagnostics}: a response can name the right reaction family and still receive no credit if the final drawn structure is not uniquely correct.

\paragraph{Partial credit depends on explicit, checkable commitments.}
The expert notes often reward model behavior that ordinary automatic metrics would miss.
For IPhO drawing tasks, GPT-5 was sometimes credited when it could not literally draw a figure but gave an unambiguous verbal description of the correct picture.
Conversely, chemistry answers with vague structural prose were often marked zero, even when the prose contained relevant terms, because the grader could not reconstruct a unique molecule.
The IChO notes also distinguish valid error-forward reasoning from invalid carryover: if a later step is consistent with an earlier explicitly drawn wrong structure, partial credit may be possible; if the earlier structure is ambiguous or chemically impossible, later prose is not enough.
This distinction became a strict rule in our judge prompt: score only what is explicitly present, do not infer missing structures, and apply error-forward only when the candidate's intermediate is concrete and chemically coherent.

\paragraph{Human comments sharpen judge calibration.}
The archived IChO grading files include cases where experts commented on both the model answer and its automatic score.
They identify demanding distinctions such as an incorrect carbon skeleton, a wrong nucleophile approach, or a correct net reaction expressed in the wrong requested form.
Accordingly, the judge pipeline uses rubric gates for answer format, structure specificity, and subpart-by-subpart score allocation.
These gates turn expert observations into an explicit, auditable checklist for structure-heavy chemistry.

\subsection{Ability Axis Glossary}
\label{sec:append-ability}
The expert comments were then abstracted into reusable ability axes.
The axes are intentionally concrete: each one corresponds to a type of credit or deduction that appeared repeatedly in medalist grading.
\begin{itemize}[leftmargin=*, noitemsep, topsep=2pt]
\item \textbf{Phys: Symbolic Deriv.} --- multi-step algebraic manipulation and equation-based reasoning.
\item \textbf{Phys: Numeric+Units} --- numerical accuracy with unit/scale discipline (dimensional checks, sanity bounds).
\item \textbf{Phys: Approx.\ \& Limits} --- correct approximations and limiting-case validation.
\item \textbf{Phys: Global Coherence} --- maintain consistent assumptions/variables across coupled subparts (regime tracking).
\item \textbf{Phys: Graph Reading} --- extract trends/values from curves and use them downstream.
\item \textbf{Shared: Visual Grounding} --- bind non-text evidence to symbols/claims (diagrams/structures/figures; ``read-the-figure'' reliability).
\item \textbf{Chem: Quant Modeling} --- quantitative physical chemistry (equilibria/thermo/electro) under algebraic constraints.
\item \textbf{Chem: Structure+Stereo} --- structure fidelity and stereochemical correctness (connectivity, configuration).
\item \textbf{Chem: Analytical} --- interpret analytical evidence (e.g., spectroscopy/assays) into chemically valid conclusions.
\item \textbf{Bio: Knowledge} --- biology knowledge and reading comprehension in constrained formats (mechanisms/terminology).
\item \textbf{Bio: Data/Exp} --- reasoning from experimental data (figures/tables, controls, statistical cues, causality).
\item \textbf{Shared: Low Halluc.} --- calibration/faithfulness proxy: lower rate of confidently-wrong outputs when confidence is available.
\end{itemize}

\subsection{Domain-specific ability profiles}
\label{app:domain_radars}
\begin{figure}[t]
  \centering
  \setlength{\abovecaptionskip}{0.1cm}
  \includegraphics[width=\linewidth]{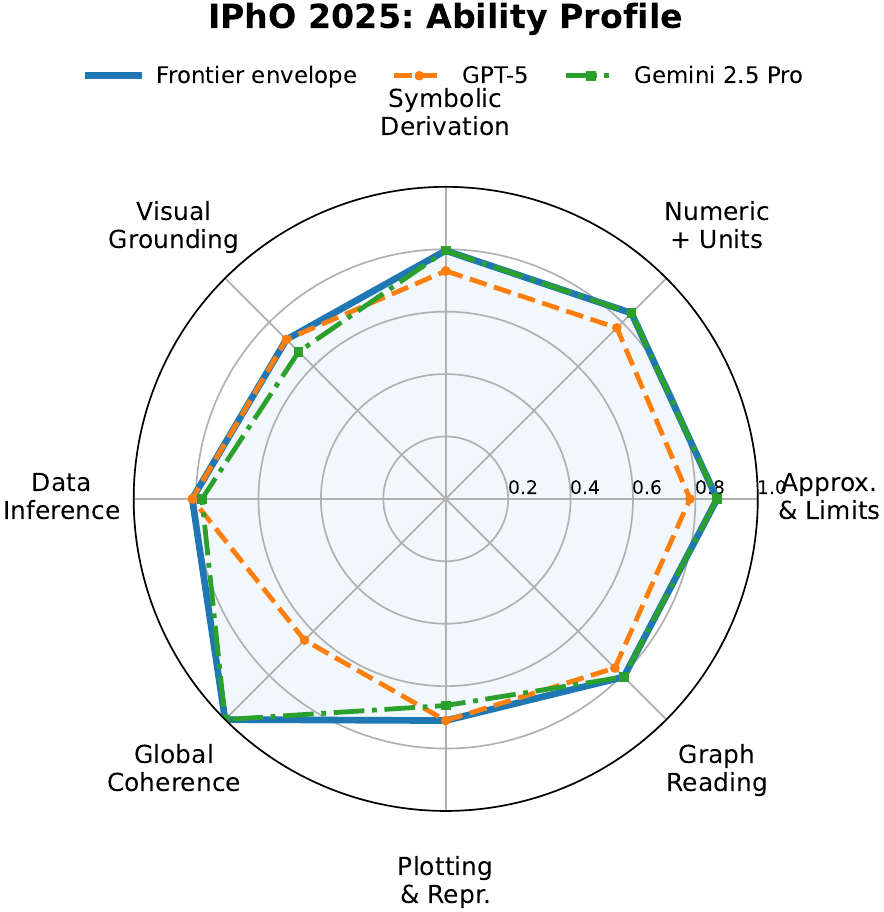}
  \caption{\textbf{IPhO 2025.} Physics abilities follow our rubric: symbolic derivation, numerical computation with units, approximation/limiting cases, graph reading, plotting/representation, global coherence across regimes, data inference from empirical relations, and visual grounding.}
  \label{fig:radar_ipho25}
\vspace{-0.3cm}
\end{figure}

\begin{figure}[t]
  \centering
  \setlength{\abovecaptionskip}{0cm}
  \includegraphics[width=\linewidth]{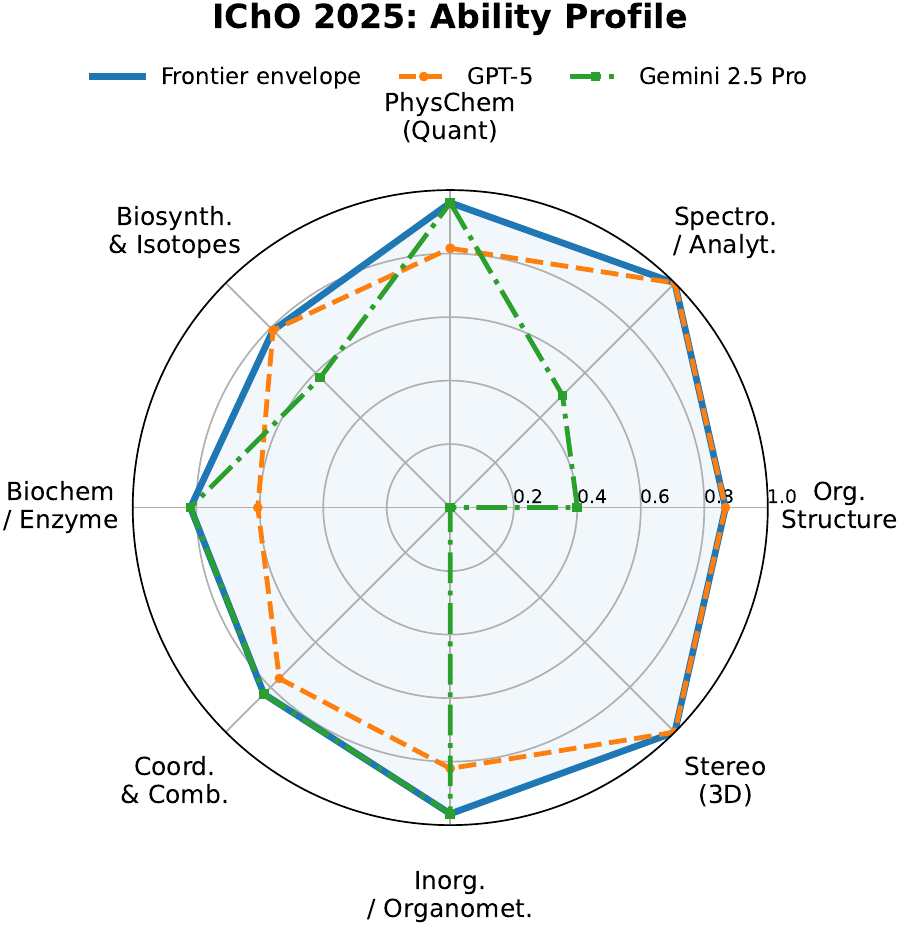}
  \caption{\textbf{IChO 2025.} Chemistry abilities use the human-coded decomposition: quantitative physical chemistry; spectroscopy/analytical reasoning; organic structure depiction; stereochemistry; inorganic/organometallic reasoning; coordination/combinatorics; biochemistry/enzyme reasoning; and biosynthesis/isotopic tracing.}
  \label{fig:radar_icho25}
  \vspace{-0.5cm}
\end{figure}

\begin{figure}[t]
  \centering
  \setlength{\abovecaptionskip}{0.1cm}
  \includegraphics[width=\linewidth]{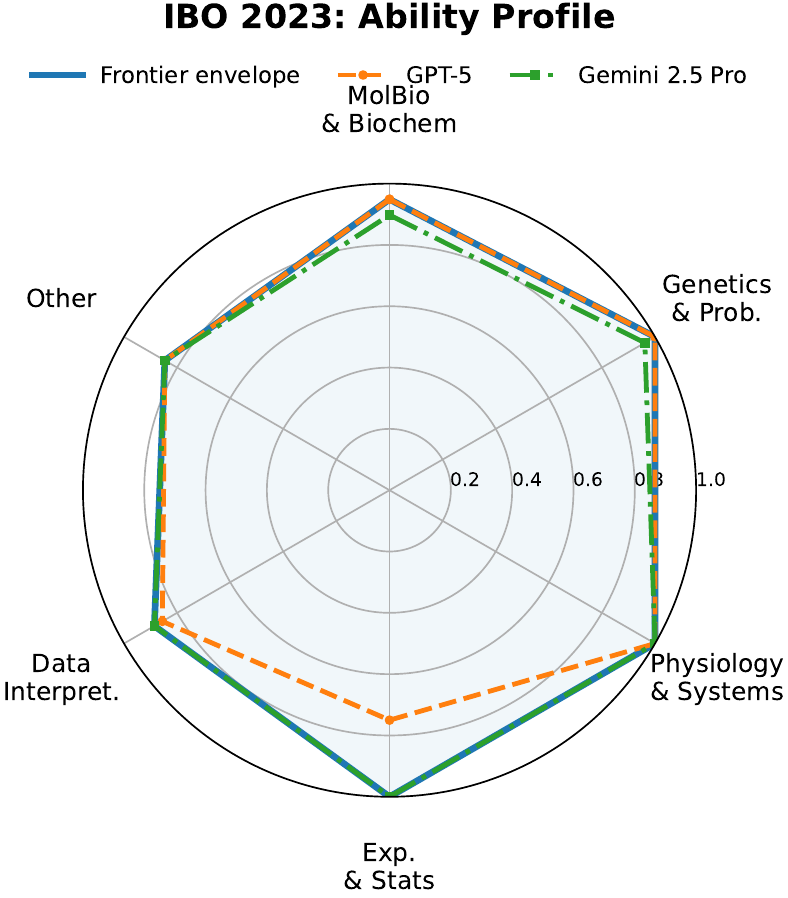}
  \caption{\textbf{IBO 2023.} Biology abilities are summarized into observed question categories derived from model outputs: molecular biology/biochemistry, genetics/probability, physiology/systems, experimental design/statistics, data/figure interpretation, and a ``other'' bucket.}
  \label{fig:radar_ibo23}
\vspace{-0.7cm}
\end{figure}

Figure~\ref{fig:radar_ipho25}, Figure~\ref{fig:radar_icho25}, and Figure~\ref{fig:radar_ibo23} summarize within-domain ability profiles induced from the expert-note taxonomy. Each radar overlays a \emph{frontier envelope} (per-axis best among the compared frontier models), GPT-5, and Gemini 2.5 Pro. Because different axes may be supplied by different systems, the frontier envelope is a diagnostic upper bound rather than the profile of a single model. Axes are domain-native (Physics/Chemistry/Biology) and are \emph{not} directly comparable across figures.

\noindent\textbf{IPhO 25 notes.} Both models are strongest on local symbolic manipulation and routine numerics; remaining errors concentrate in \emph{global coherence} (propagating assumptions consistently across long solutions) and \emph{visual grounding} (extracting non-textual cues), with noticeable variance across tasks requiring regime switching.

\noindent\textbf{IChO 25 notes.} Gemini 2.5 Pro tends to be more uniformly strong on quantitative and analytical axes (physical chemistry and spectroscopy), while GPT-5 exhibits higher dispersion: it can solve isolated substeps but is less consistent when multiple constraints (structure + stereochemistry + quantitative checks) must be satisfied simultaneously.

\noindent\textbf{IBO 23 notes.} Frontier models reach near-ceiling performance on knowledge-centric multiple-choice biology (molecular biology and genetics), but are noticeably weaker on evidence-centric reasoning (statistics/design and data/figure interpretation), consistent with broader multimodal grounding limitations. Together, the three profiles show that remaining weaknesses are concentrated in evidence binding, representational precision, and long-horizon consistency rather than simple recall.

\section{Prompt Templates}
\label{app:prompts}
The benchmark uses generated prompts at the turn level; below are the reusable templates and the sealed 2026 routing invariant.
Problem-specific fields are filled from the released JSON rows. Text-only routes contain no image blocks or remote image URLs, whereas vision routes in the sealed 2026 runs receive the exact original question PNG bytes as API image blocks.

\begin{tcolorbox}[
  title={Benchmark digitization prompt},
  colback=gray!3,
  colframe=black!35,
  boxrule=0.6pt,
  arc=1.5mm,
  left=1.5mm,right=1.5mm,top=1mm,bottom=1mm
]
\footnotesize
Convert the provided competition problem, rendered pages, figures, official answer, detailed solution, and rubric into canonical text.
Preserve all numerical values, labels, tables, figure relationships, chemical structures, diagrams, and point allocations.
If a figure is needed for solving, describe it explicitly enough that a text-only solver can use it.
Do not introduce outside facts, do not omit subparts, and do not include remote URLs or image-data strings in solver-visible fields.
Return structured JSON with question text, figure descriptions, official answer, detailed solution, rubric, source caveats, and completeness notes.
\end{tcolorbox}

\begin{tcolorbox}[
  title={Sealed 2026 question-only staticization prompt for text routes},
  colback=gray!3,
  colframe=black!35,
  boxrule=0.6pt,
  arc=1.5mm,
  left=1.5mm,right=1.5mm,top=1mm,bottom=1mm
]
\footnotesize
Convert only the supplied official question text and question-page images into answer-blind canonical problem text for a text-only solver.
Preserve every visible number, label, table, diagram relation, chemical structure, and requested answer format needed to state the question.
Do not use or infer any official answer, solution, marking rubric, or solution-page content.
Return only the frozen question-side representation and provenance/completeness fields; include no URLs or image-data strings.
\end{tcolorbox}

\begin{tcolorbox}[
  title={Interleaved solver prompt for open-ended items},
  colback=gray!3,
  colframe=black!35,
  boxrule=0.6pt,
  arc=1.5mm,
  left=1.5mm,right=1.5mm,top=1mm,bottom=1mm
]
\footnotesize
You are an expert olympiad contestant in the relevant scientific domain.
Solve the current subpart using only the provided staticized problem text, prior subpart context from the same sequence, and the official answer-format instruction.
Show enough reasoning for the grader to identify partial-credit steps, keep variables and assumptions consistent with earlier turns, and make the final answer explicit.
When a structure, diagram, or table entry is required, describe the complete checkable object rather than only naming a reaction, concept, or qualitative trend.
\end{tcolorbox}

\begin{tcolorbox}[
  title={Sealed 2026 vision-route payload invariant},
  colback=gray!3,
  colframe=black!35,
  boxrule=0.6pt,
  arc=1.5mm,
  left=1.5mm,right=1.5mm,top=1mm,bottom=1mm
]
\footnotesize
Send the native question text and every frozen question-page PNG needed by the current linked sequence directly as image blocks in the solver API request.
The image bytes must match the official question-page render manifest; do not substitute a caption or description generated by another multimodal model.
Never attach official answers, solutions, rubrics, or solution-page images to the solver request.
\end{tcolorbox}

\begin{tcolorbox}[
  title={IBO objective-key prompt},
  colback=gray!3,
  colframe=black!35,
  boxrule=0.6pt,
  arc=1.5mm,
  left=1.5mm,right=1.5mm,top=1mm,bottom=1mm
]
\footnotesize
Answer the biology competition question based solely on the provided information.
For true/false questions output T or F; for choice questions output the requested letter or number; for numerical blanks output the requested rounded value.
Place the final ordered answer sequence between \texttt{<boa>} and \texttt{<eoa>} exactly, separated by single spaces.
Explanations may appear outside the tags, but the text inside the tags must contain only the final answer tokens.
\end{tcolorbox}

\begin{tcolorbox}[
  title={Rubric-faithful judge prompt},
  colback=gray!3,
  colframe=black!35,
  boxrule=0.6pt,
  arc=1.5mm,
  left=1.5mm,right=1.5mm,top=1mm,bottom=1mm
]
\footnotesize
You are a strict olympiad grader.
Given the problem, official answer, detailed solution, point rubric, and candidate response, assign partial credit exactly according to the rubric.
Score only explicit, checkable claims in the candidate response.
Do not infer missing structures, diagrams, numerical values, or reaction intermediates.
Apply error-forward only when the candidate has made a concrete intermediate commitment that is internally coherent and the later reasoning follows from it.
When original question or solution page images are attached to the judge packet, treat those images and the provenance-labeled official rubric as authoritative evidence.
Return JSON with \texttt{score}, \texttt{max\_score}, subcriterion scores, concise evidence, and any uncertainty or escalation flag.
\end{tcolorbox}

\begin{tcolorbox}[
  title={Subfield and modality classification prompt},
  colback=gray!3,
  colframe=black!35,
  boxrule=0.6pt,
  arc=1.5mm,
  left=1.5mm,right=1.5mm,top=1mm,bottom=1mm
]
\footnotesize
Classify the benchmark turn using only the staticized question, official answer, and rubric.
Return JSON fields for scientific domain, subfield, skill type, whether visual evidence is required, whether chemical/biological structures are required, whether quantitative calculation is required, and a short rationale.
Do not look at model answers and do not change any benchmark content.
\end{tcolorbox}

\section{2026 Evaluation and Calibration Audit}
\label{app:fresh-audit}

\paragraph{Evaluation completeness.}
The four 2026 runs contain 23 IPhO, 68 IChO, 28 NBPhO, and 34 USAPhO linked turns, respectively.
All 2,142 solver terminals and all 56 model--contest aggregates are present.
For every vision-route attempt, the audit reconstructs the API payload and verifies each direct PNG block against the frozen official-page SHA-256; every text-route attempt has zero image blocks and a recomputed question-only prompt hash.
Every non-deterministic judge request binds the fixed Gemini 3.5 Flash route, exact item rubric, criterion identifiers, and maxima.

\paragraph{Route-level macro scores and input assignments.}
\setcounter{savedtablecounter}{\value{table}}
\setcounter{table}{6}
\begin{table}[t]
\centering
\scriptsize
\setlength{\tabcolsep}{3.2pt}
\renewcommand{\arraystretch}{1.02}
\begin{tabular}{l c r r}
\toprule
\textbf{Model} & \textbf{Input} & \textbf{Macro} & \textbf{95\% CI} \\
\midrule
Gemini 3 Flash & V & 85.56 & [79.24, 91.66] \\
Seed 2.0 Pro & V & 71.03 & [63.67, 78.85] \\
Gemini 3.1 Pro & V & \RankThree{88.47} & [83.80, 93.18] \\
GPT-5.4 & V & 68.79 & [61.86, 75.78] \\
GLM-5.1 & T & 86.89 & [81.13, 92.05] \\
Qwen3.6-Plus & V & 82.63 & [75.55, 89.74] \\
MiniMax M2.7 & T & 69.61 & [62.01, 77.48] \\
Claude Opus 4.7 & V & 79.86 & [73.19, 86.37] \\
MiMo V2.5 Pro & T & 63.81 & [54.46, 72.66] \\
GPT-5.5 & V & \RankOne{93.31} & [90.95, 95.88] \\
DeepSeekV4-Pro & T & 87.76 & [82.86, 92.43] \\
DeepSeekV4-Flash & T & 86.62 & [80.97, 91.99] \\
Gemini3.5-Flash & V & \RankTwo{89.54} & [86.12, 93.19] \\
Qwen3.7-Max & T & 85.16 & [79.82, 90.54] \\
\bottomrule
\end{tabular}
\caption{Four-contest normalized macro scores and sequence-cluster bootstrap intervals. V routes receive original question PNGs directly; T routes receive answer-blind question-only staticizations and no images. The interval resamples whole problem sequences within each contest and captures item/sequence sampling uncertainty only, not generation or judge variance.}
\label{tab:fresh-2026-ci}
\end{table}
\setcounter{table}{\value{savedtablecounter}}

The four-contest evaluation contains 153 linked turns grouped into 28 whole problem sequences, yielding 2,142 sealed model--turn terminals and complete aggregates for all 14 routes. Payload audits verify that every stored V request includes direct PNG data-image blocks whose bytes match the frozen official question pages, while every T request contains zero image blocks and exactly the frozen question-only staticization. All four contest evaluations use the same route roster, modality assignments, score maxima, fixed Gemini 3.5 Flash judge route, and frozen rubric criteria.

The 10,000 paired resamples use seed 20260829 and sample whole problem sequences: 3 IPhO, 9 IChO, 10 NBPhO, and 6 USAPhO problems, with all linked subparts moving together. Each contest score is normalized by its explicit maximum before the four percentages are equally averaged, and no missing aggregate is imputed. Intervals overlap substantially for several neighboring systems, so small point-ranking differences should not be read as statistically resolved. Exact first-availability days and training cutoffs are unavailable for the evaluated routes; same-month status is therefore unknown. Official day-level problem-release evidence is available for NBPhO (25 April 2026) and USAPhO (20 May 2026), whereas the official IPhO and IChO pages display no publication date.

\begin{table}[t]
\centering
\scriptsize
\setlength{\tabcolsep}{2.5pt}
\resizebox{\columnwidth}{!}{%
\begin{tabular}{l r r r r l}
\toprule
\textbf{Domain} & \textbf{Native} & \textbf{Description} & \textbf{Blind} & \textbf{$\Delta$ N--D} & \textbf{95\% CI} \\
\midrule
Physics & 78.82 & 81.57 & 55.47 & $-2.75$ & [$-6.62$, 1.15] \\
Chemistry & 53.72 & 48.70 & 17.73 & $+5.01$ & [0.57, 9.56] \\
Biology & 85.79 & 84.99 & 76.50 & $+0.80$ & [$-2.13$, 3.74] \\
\bottomrule
\end{tabular}%
}
\caption{Domain split of the controlled modality ablation. Arm entries are mean normalized scores in percent; $\Delta$ N--D is native image minus faithful description in percentage points, with paired-bootstrap intervals.}
\label{tab:modality-domain-split}
\vspace{-0.3cm}
\end{table}

The transport audit confirms delivery of every scheduled native image block across all retained API requests, passes visible-image-use checks for every retained native response, and finds no image blocks or hidden URLs in either text arm. These checks confirm that the three conditions differ in visual evidence as designed rather than through unintended route leakage.

\paragraph{Frozen-answer cross-family judge study.}
\begin{table}[t]
\centering
\scriptsize
\setlength{\tabcolsep}{2.2pt}
\renewcommand{\arraystretch}{1.02}
\begin{tabular}{l c r r r}
\toprule
\textbf{Judge} & \textbf{Input} & \textbf{IPhO (/30)} & \textbf{IChO (/60)} & \textbf{MAE (pp)} \\
\midrule
Gemini 3.5 Flash & V & $29.750\pm0.000$ & $56.011\pm0.516$ & 1.582 \\
Gemini 3.1 Pro & V & $28.750\pm1.000$ & $56.456\pm0.097$ & 0.866 \\
Claude Opus 4.7 & V & $29.083\pm1.155$ & $56.738\pm0.508$ & 2.323 \\
GPT-5.5 & V & $28.250\pm0.866$ & $57.099\pm0.192$ & 0.733 \\
Qwen3.7-Max & T & $27.750\pm1.000$ & $54.442\pm0.815$ & 1.691 \\
DeepSeek V4 Pro & T & $28.083\pm0.577$ & $54.761\pm1.568$ & 1.672 \\
\bottomrule
\end{tabular}
\caption{Three-repeat judging of the same frozen Gemini 3.5 Flash answers from IPhO and IChO 2026. Contest entries are mean $\pm$ sample SD; MAE is the overall within-judge normalized repeat MAE in percentage points. V and T denote the problem representation: V uses original official question and available solution/marking PNGs, whereas T replaces problem images with the frozen question-side staticization and sends zero image blocks; both modes retain the same candidate-answer, official-solution, and rubric packet.}
\label{tab:cross-family-judge-stability}
\vspace{-0.3cm}
\end{table}

Strict replay covers all 1,638 effective final judge terminals. Gemini 3.1 Pro uses one same-configuration recovery cell, and the DeepSeek row uses one unified thinking-enabled 65k-token configuration over all 273 cells; only these matched-configuration results enter the table. The four V rows define the primary cross-family comparison, while the two T rows are modality-sensitivity conditions and are not pooled into V--V agreement.

\paragraph{Archived calibration replay.}
The structured replay finds 1,050 stored judgment rows over 525 distinct expert rows, with no missing or errored judgment. Expert-agreement metrics use 1,048 scope-aligned pairs. One IChO expert label for each judge scores the full multipart Q2.3 group, whereas the corresponding judge row scores only Q2.3c on its six-point rubric; these two records address different units and therefore do not enter item-level agreement metrics.

\paragraph{Structured expert feedback.}
The deterministic export contains 525 provenance-addressed expert-score rows: all 325 IChO rows and 85 of 200 IPhO rows retain nonempty comments, while the remaining 115 IPhO rows explicitly record a null comment; each row also preserves the two archived calibrated-judge records when available.
The archived sources contain no human-authored error tags, evidence spans, uncertainty labels, or conflict labels; the corresponding fields remain null and are not synthesized automatically.
The resulting schema supports item-level analysis of judge--expert discrepancies while retaining exact provenance to the original grading records.

\begin{table}[t]
\centering
\scriptsize
\setlength{\tabcolsep}{3.2pt}
\begin{tabular}{l l r r r r}
\toprule
\textbf{Judge} & \textbf{Exam} & \textbf{$n$} & \textbf{Raw MAE} & \textbf{Norm. MAE} & \textbf{$r$} \\
\midrule
Gemini 3.1 Pro & IPhO 25 & 200 & 0.066 & 0.092 & 0.833 \\
Gemini 3.1 Pro & IChO 25 & 324 & 0.303 & 0.075 & 0.909 \\
Gemini 3.5 Flash & IPhO 25 & 200 & 0.069 & 0.097 & 0.829 \\
Gemini 3.5 Flash & IChO 25 & 324 & 0.537 & 0.138 & 0.868 \\
\bottomrule
\end{tabular}
\caption{Item-level replay of the two calibrated judges against expert rows under the frozen scoring protocol. Raw and normalized metrics use the same 324 scope-aligned IChO pairs; the whole-group Q2.3 expert label is outside this item-level estimand. Cross-family robustness is evaluated separately in Table~\ref{tab:cross-family-judge-stability}.}
\label{tab:archived-judge-replay}
\vspace{-0.3cm}
\end{table}

\section{Additional Temporal and Judge Sensitivity Checks}

The following analyses test two complementary sensitivities: the temporal probe tests direct answer memorization, and the IChO subset stress-tests structure-aware grading. They supplement the repeated-run and cross-family experiments by targeting distinct aspects of evaluation reliability.

\paragraph{Exploratory temporal probe.}
Across three models and four seeds, 69/72 isomorphic numeric perturbations elicited the perturbed target, with 0/72 stale original-answer substitutions; a strict no-question control produced 0/72 recalls.
This small behavioral probe finds no direct memorization signal, but it cannot establish absence of training exposure and does not change our public-benchmark interpretation of Table~\ref{tab:latest-all-benchmarks}.

\paragraph{Structure-heavy judge stress test.}
On a 50-row IChO structure/stereochemistry subset scored by each of four image-aware judges, MAE against expert scores ranges from 0.66 to 1.24 points; the numbers of rows over-credited by at least 0.5 point are 5, 8, 9, and 21 out of 50 across the four judges.
Across this demanding subset, all four judges remain close to expert scores; multi-judge agreement or expert adjudication provides the strongest safeguard for the lowest-confidence structural cases.

\section{Competition Sources}
\label{app:competitions}

\textsc{ScienceArena} is built from thirteen olympiad-style scientific competitions spanning physics, chemistry, and biology, with explicit source provenance. We record the \emph{public release time} of official problem materials separately from the contest year for descriptive temporal screening. When an exact official publication date or evaluated- route availability date is unavailable, it is marked unknown rather than treated as a strict time-release holdout; no release timestamp by itself proves absence from training data. Table~\ref{tab:science_arena_sources} summarizes the benchmark sources and rubric status.

\paragraph{International Physics Olympiad (IPhO 2025).}
We use the official IPhO 2025 examination page and its released questions and answer sheets \citep{ipho2025_official_questions}. IPhO exams consist of theory and experimental components; our benchmark focuses on the \emph{theory} problems, since experimental tasks require physical apparatus and proctoring conditions that are not directly reproducible for LLM evaluation.

\paragraph{International Chemistry Olympiad (IChO 2025).} 
We source IChO 2025 materials from the official English examination-paper release \citep{icho2025_problems_page}. Similar to IPhO, IChO includes theory and laboratory components; we evaluate models on the \emph{theory} portion to ensure a well-defined, reproducible setting with rubric-faithful scoring across all models.

\paragraph{International Physics and Chemistry Olympiads (IPhO/IChO 2026).}
The fresh international tracks use the official IPhO questions page and the official IChO problems-and-solutions page \citep{ipho2026_official_questions,icho2026_official_problems}.
The 56th IPhO ran from 4--12 July 2026 and held its theoretical examination on 8 July; the 58th IChO ran from 10--19 July 2026 and held its theoretical examination on 15 July \citep{ipho2026_official_program,icho2026_official_program}. The official hosts expose the evaluated examination files with July 2026 file metadata, so the tables record the month-level release as 26-07; this does not assert an exact first-publication day.

\paragraph{International Biology Olympiad (IBO 2023).}
IBO provides an official archive of examination papers for past competitions \citep{ibo_papers_archive}; the archived 2023 materials used here were released in August 2025. In Science Arena, IBO 2023 serves as an \emph{objectively scorable} biology track (e.g., multiple-choice or otherwise unambiguously keyed questions), enabling automatic evaluation without judge-model mediation. When reporting medal-calibrated comparisons, we reference the official IBO 2023 results report \citep{ibo2023_final_results}.

\paragraph{China Chemistry Olympiad (CChO 2025).}
For CChO 2025, we use publicly released final-round materials (problem statements and marking guidance) distributed as a consolidated PDF \citep{ccho2025_final_theory_paper_public_pdf}. When normalizing scores and aligning evaluation to official scoring practice, we additionally cite the Chinese Chemical Society document describing the theory score conversion weights \citep{ccho2025_theory_weighting_chemsoc}. As with other laboratory-inclusive competitions, our primary benchmark setting targets the theory component for reproducibility.

\paragraph{China Physics Olympiad (CPhO 2025).}
For CPhO 2025, we use publicly released final-round theory materials and the associated marking scheme \citep{cpho2025_final_theory_paper_public_pdf,cpho2025_final_theory_solution_marking_public_pdf}. Where relevant to describing the competition formally (e.g., structure, governance, or official rules), we also cite the published competition charter \citep{cpho_charter_2025_revision}. We evaluate the theory component as the reproducible portion of the exam; experimental components and their marking documents can be referenced separately when needed \citep{cpho2025_final_practical_paper_public_pdf,cpho2025_final_practical_answer_key_public_pdf}.

\paragraph{Additional regional and national sources.}
The expanded suite also includes APhO 2025, EuPhO 2025, NBPhO 2026, USAPhO 2026, INChO 2026, and USNCO 2025. NBPhO problem media are dated 25 April 2026 on the official host, while the official USAPhO news page links the exam and solution under 20 May 2026 \citep{nbpho2026_official_event,nbpho2026_media_metadata,usapho2026_official_news}. These sources are normalized through the same digitization and expert-audit pipeline described in Section~\ref{sec:2}; source caveats and rubric are kept in the released metadata.

\paragraph{Notes on scope and reproducibility.}
All open-ended competitions are high-stakes, rubric-driven assessments with well-defined problem sets and partial-credit grading, while IBO is objectively keyed. \textsc{ScienceArena} prioritizes \emph{publicly released} materials for reproducibility and transparent sourcing, and it uses theory-only evaluation where experimental replication would otherwise confound model comparisons. Each digitized item preserves its source examination, official answer, and grading rules, making the evaluation packet auditable. Frozen item definitions and score maxima are shared across model routes, ensuring every system is evaluated against the same scientific target.




\end{document}